%% file: main.tex
\documentclass[10pt,twocolumn,letterpaper]{article}

\usepackage[pagenumbers]{cvpr} % To force page numbers, e.g. for an arXiv version
\usepackage{tabularray}
\usepackage{tabularx}


\definecolor{cvprblue}{rgb}{0.21,0.49,0.74}
\usepackage[pagebackref,breaklinks,colorlinks,allcolors=cvprblue]{hyperref}
\usepackage{tcolorbox}
\tcbuselibrary{breakable}
\title{InsightVision: A Comprehensive, Multi-Level Chinese-based Benchmark for Evaluating Implicit Visual Semantics in Large Vision Language Models}
\author{Xiaofei Yin\thanks{}\\
Ant Security Lab, Ant Group\\
Shanghai, China\\
{\tt\small yinxiaofei.yxf@antgroup.com}
\and
Yijie Hong\\
Shanghai Jiaotong University\\
Shanghai, China\\
{\tt\small 1656125037@sjtu.edu.cn}
\and
Ya Guo\\
Ant Security Lab, Ant Group\\
Shanghai, China\\
{\tt\small guoya.gy@antgroup.com}
\and
Yi Tu\\
Ant Security Lab, Ant Group\\
Shanghai, China\\
{\tt\small qianyi.ty@antgroup.com}
\and
Weiqiang Wang\\
Ant Security Lab, Ant Group\\
Hangzhou, China\\
{\tt\small weiqiang.wwq@antgroup.com}
\and
Gongshen Liu\\
Shanghai Jiaotong University\\
Shanghai, China\\
{\tt\small lgshen@sjtu.edu.cn}
\and
Huijia zhu\thanks{}\\
Ant Security Lab, Ant Group\\
Shanghai, China\\
{\tt\small huijia.zhj@antgroup.com}
}

\begin{document}
\maketitle
\input{sec/0_Abstract}    
\input{sec/1_Intro}
\input{sec/2_Related_Work}
\input{sec/3_Method}

\input{sec/4_Experiments}

\input{sec/5_Conclusion}

{
    \small
    \bibliographystyle{ieeenat_fullname}
    \bibliography{main}
}
\input{sec/appendix}


\end{document}

%% file: sec/0_Abstract.tex
\begin{abstract}
In the evolving landscape of multimodal language models, understanding the nuanced meanings conveyed through visual cues—such as satire, insult, or critique—remains a significant challenge. Existing evaluation benchmarks primarily focus on direct tasks like image captioning or are limited to a narrow set of categories, such as humor or satire, for deep semantic understanding. To address this gap, we introduce, for the first time, a comprehensive, multi-level Chinese-based benchmark designed specifically for evaluating the understanding of implicit meanings in images. This benchmark is systematically categorized into four subtasks: surface-level content understanding, symbolic meaning interpretation, background knowledge comprehension, and implicit meaning comprehension. We propose an innovative semi-automatic method for constructing datasets, adhering to established construction protocols. Using this benchmark, we evaluate 15 open-source large vision language models (LVLMs) and GPT-4o, revealing that even the best-performing model lags behind human performance by nearly 14\% in understanding implicit meaning. Our findings underscore the intrinsic challenges current LVLMs face in grasping nuanced visual semantics, highlighting significant opportunities for future research and development in this domain. We will publicly release our InsightVision dataset, code upon acceptance of the paper.
\end{abstract}

%% file: sec/1_Intro.tex
\section{Introduction}
\label{sec:intro}
\begin{figure}[t]
    \centering
    \includegraphics[width=1\linewidth]{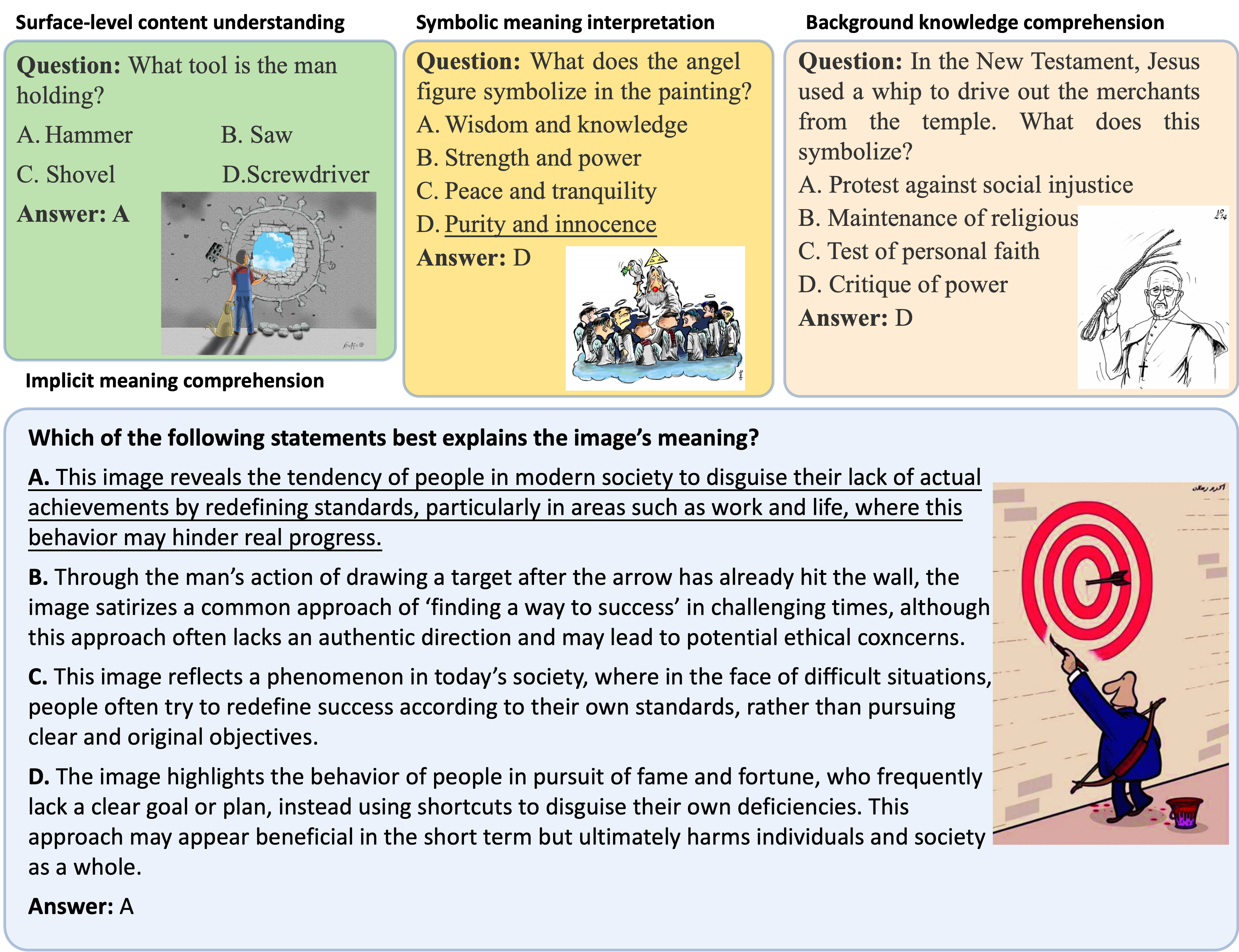}
    
    \caption{Several examples from the InsightVision dataset. Chinese questions and answers have been translated into English.}
    \label{fig:example}
\end{figure}

In the domain of multimodal language models\cite{achiam2023gpt, li2024llava,you2023ferretrefergroundgranularity}, grasping the subtle meanings conveyed through visual cues—such as sarcasm, insult, or criticism—remains a substantial challenge. Understanding the nuanced implications of images is indicative of advanced human intelligence, serving as a vital bridge between perceptual and cognitive intelligence\cite{gordon2019intermodulation,DEWIT2012665}. Many images cannot be fully comprehended by merely examining their surface content; instead, a genuine understanding requires integrating background knowledge and symbolic cues to discern the true intentions of the image's creator\cite{garner1987metacognition,wang2024browseconcentratecomprehendingmultimodal}.

While visual perception entails transforming visual signals into insightful conclusions, such as profound image semantics or subtle narrative tones, existing evaluation benchmarks often fall short of assessing these deeper levels of understanding\cite{Goyal_2017_CVPR, Hiippala_2020}. These benchmarks primarily emphasize superficial tasks, such as image captioning, with datasets like COCO and ImageNet\cite{Hudson_2019_CVPR, cai2019multi}. Such efforts inadequately capture the intricacies of symbolic meanings and implicit interpretations. Furthermore, comprehensive visual perception demands both high- and low-level understanding, whereby humans employ commonsense knowledge to interpret broad concepts before honing in on the details\cite{wang2024browseconcentratecomprehendingmultimodal,chow2023travlritdontbimodal}. Current large vision-language models (LVLMs), however, often show limitations in articulating this hierarchical understanding.

To address these challenges and bridge the gaps in existing research, we introduce InsightVision, a comprehensive Chinese-based benchmark designed for nuanced, multi-level image evaluation. The InsightVision is systematically divided into four subtasks: surface-level content understanding, background knowledge comprehension, symbolic meaning interpretation, and implicit meaning comprehension. Unlike traditional datasets, it aims to provide a more thorough evaluation of multimodal language models' ability to grasp the deep semantics underlying images. The dataset comprises over 2,500 samples, each consisting of an image accompanied by questions spanning the four dimensions. Additionally, we have developed a semi-automatic pipeline to construct high-quality dataset. Utilizing InsightVision, we evaluate the implicit understanding capabilities of 15 open-source LVLMs and GPT-4o. Our assessment reveals a substantial gap between existing LVLMs and human performance in comprehending implicit meanings. For instance, even the best-performing model lags behind humans by nearly 14\% in terms of understanding implicit implications. These findings highlight the significant challenges in this domain and underscore the substantial opportunity for improvement in developing models capable of deeply understanding visual semantics. We have publicly released our annotations, code, and model results. We will publicly release our InsightVision dataset, code upon acceptance of the paper.

%第1、2段需要找参考文献
 

%% file: sec/2_Related_Work.tex
\section{Related Work}
\label{sec:Related Work}
\subsection{Large vision language model}
Vision-language models\cite{li2023blip,li2024llava,bai2023qwen,lu2024deepseekvlrealworldvisionlanguageunderstanding, alayrac2022flamingo,sun2024generativemultimodalmodelsincontext}have achieved remarkable advancements within the realm of multimodal intelligence. By amalgamating large language models\cite{ray2023chatgpt,achiam2023gpt,anil2023palm,touvron2023llama2openfoundation,touvron2023llamaopenefficientfoundation} with visual content, LVLMs effectively manage intricate visual and linguistic inputs, thereby executing a variety of tasks ranging from visual description to logical reasoning. Flamingo\cite{alayrac2022flamingo} and OpenFlamingo\cite{awadalla2023openflamingoopensourceframeworktraining} models incorporate visual feature processing modules into the internal strata of language models using gated cross-attention, thereby propelling the profound integration of visual data within LLMs. CLIP\cite{radford2021learning,sun2023evaclipimprovedtrainingtechniques} utilizes contrastive learning to harmonize image and text modalities and is trained on extensive, noisy web-derived image-text pairs. By integrating modules such as QFormer\cite{li2023blip} and MLP\cite{liu2024visual}, previous works\cite{bai2023qwen, dai2023instructblipgeneralpurposevisionlanguagemodels,Liu_2024_CVPR} facilitate a collaborative comprehension between visual encoders and large language models (LLMs) of multimodal inputs. LLaVA\cite{liu2024visual} stands out for its pioneering use of GPT-generated instruction-following data to amplify LVLMs' responsiveness to visual instructions. A plethora of powerful LVLM APIs, including GPT-4o\cite{achiam2023gpt} and Qwen-VL-max\cite{bai2023qwen}, are now available. Through a rigorous evaluation of these models based on our proposed benchmark, we offer insightful perspectives into the ongoing research surrounding LVLMs.
\subsection{Vision Language Benchmarks} A rapidly expanding suite of multimodal benchmarks now rigorously evaluates the capabilities of LVLMs. Established benchmarks, including COCO Caption \cite{chen2015microsoftcococaptionsdata}, VQAv2 \cite{Goyal_2017_CVPR}, and GQA \cite{Hudson_2019_CVPR}, predominantly center on image description and question-answering tasks, employing metrics such as BLEU, CIDEr, and accuracy to gauge performance. Yet, as LVLMs advance, these traditional datasets have become insufficient for fully capturing the breadth of model capabilities. In response, researchers have developed more comprehensive evaluation frameworks that test a wider range of competencies, encompassing perceptual and cognitive skills \cite{fu2024mmecomprehensiveevaluationbenchmark}, spatial-temporal reasoning \cite{li2023seedbenchbenchmarkingmultimodalllms}, and relational understanding \cite{liu2025mmbench}. For instance, MMMU \cite{Yue_2024_CVPR} curates data from college-level textbooks and lecture materials, challenging models to demonstrate expertise across six academic disciplines. Similarly, CMMU \cite{he2024cmmubenchmarkchinesemultimodal} gathers questions from primary through high school curricula to assess foundational knowledge within the Chinese educational context. Nevertheless, these benchmarks largely remain focused on basic visual tasks, without adequately addressing the complexity of multimodal understanding. This paper introduces a benchmark tailored to evaluate deep semantic comprehension of images, specifically within a Chinese cultural framework.
\subsection{Image implicit meaning comprehension}
Image implicit meaning comprehension has become an important research focus for contemporary LVLMs, especially in handling images that convey complex emotions, cultural symbolism, and social critique. Existing evaluation datasets primarily test the models' linear visual reasoning abilities, such as visual question answering for surface-level content\cite{Hudson_2019_CVPR}. However, several works \cite{cai2019multi, machajdik2010affective} have demonstrated that LVLMs’ capabilities go beyond understanding surface-level meanings. Recent works\cite{yang2024largemultimodalmodelsuncover, liu2024iibenchimageimplicationunderstanding} highlight the limitations of current models when it comes to processing nonlinear narratives and understanding cultural contexts. For example, the most relevant prior work, DEEPEVAL\cite{yang2024largemultimodalmodelsuncover}, introduces three core tasks and shows that while the most advanced models achieve near-human performance on basic visual description tasks, they still perform poorly on tasks that involve understanding implicit semantics such as social background and satire. This paper provides a more comprehensive Chinese understanding benchmark, which, compared to the six categories in DeepEval, expands to include more thematic categories, with a total of 13 major categories and 41 subcategories (Figure \ref{fig:categories}), and offers more detailed testing across four dimensions of model performance.

%% file: sec/3_Method.tex
\section{Dataset and task overview}
\label{sec:Method}

\begin{table}
\centering
\begin{tblr}{
  cell{2}{2} = {c},
  cell{3}{2} = {c},
  cell{4}{2} = {c},
  cell{5}{2} = {c},
  cell{6}{2} = {c},
  hline{1,7} = {-}{0.08em},
  hline{2} = {-}{0.05em},
}
Image Amount                         & 2500  \\
QA Amount                           & 16220 \\
Surface-level Content Understanding & 5713  \\
Symbolic Meaning Interpretation     & 4649  \\
Background Knowledge Comprehension  & 3548  \\
Implicit Meaning Comprehension      & 2310  
\end{tblr}
\caption{Statistics of InsightVision dataset.}
\label{tab:dataset}
\end{table}

\begin{figure}[t]
    \centering
    \includegraphics[width=1\linewidth]{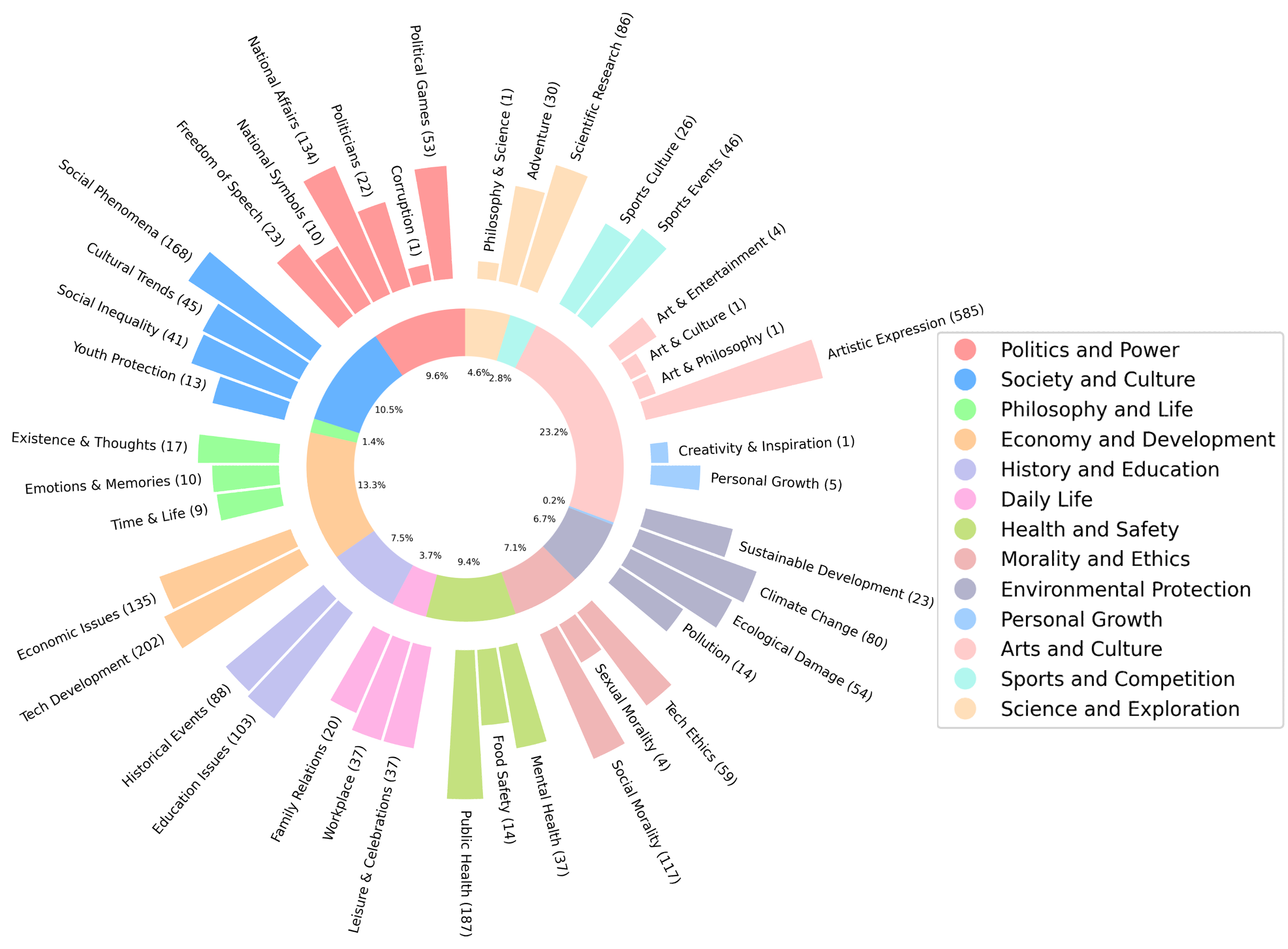}
    
    \caption{Data distribution of major categories and subcategories in InsightVision.}
    \label{fig:categories}
\end{figure}
% 这里需要加一个表1和图1

InsightVision, a comprehensive Chinese dataset, has been meticulously developed to assess the proficiency of LVLMs in deciphering nuanced and implicit meanings within visual content. This dataset encompasses 2,500 carefully curated samples, each comprising an image coupled with a set of choice questions. These questions are strategically designed to evaluate four distinct dimensions: surface-level content understanding, symbolic meaning interpretation, background knowledge comprehension, and implicit meaning comprehension.

The structure of InsightVision reflects the complex cognitive process involved in image interpretation, where models are required to first comprehend the surface visual content, then integrate extensive background knowledge and symbolic interpretations to ultimately infer the implicit meaning. To facilitate quantitative evaluation, we have crafted one or more single-choice questions for each dimension, testing the model's understanding across various levels of complexity. Each question presents an image, a query, and four answer options, with only one correct answer and three carefully designed distractors.

This holistic design in dataset construction allows for a robust evaluation of LVLMs' capabilities in processing visual information beyond mere surface-level recognition, delving into deeper levels of contextual and cultural understanding. Table \ref{tab:dataset} provides detailed textual statistics of the dataset, while Figure \ref{fig:example} illustrates some representative examples, demonstrating the dataset's comprehensive nature. %the comprehensive nature of the InsightVision dataset and its potential for advancing research in visual-language understanding within AI systems.

The four primary subtasks in our evaluation framework are:

\textbf{Surface-level content understanding}: This subtask assesses the model's ability to accurately identify and describe visual details present in the image. It serves as a foundation for more complex interpretations and ensures that the model can process basic visual information effectively.

\textbf{Symbolic meaning interpretation}. This subtask evaluates the model's capacity to understand the symbolic or metaphorical meanings conveyed by the image content. It tests the model's ability to move beyond literal interpretation and grasp deeper, culturally-informed meanings.

\textbf{Background knowledge comprehension}. This subtask evaluates the model's ability to leverage relevant background knowledge necessary for understanding the image content. It examines the model's capacity to integrate external information and context with visual cues.  %This subtask tests the model's ability to leverage relevant contextual and cultural knowledge to fully comprehend the image's content. It assesses whether the model can draw upon a broad knowledge base to enhance its understanding of the visual input.

\textbf{Implicit meaning comprehension}. The final subtask examines the model's proficiency in grasping the overall implicit message or subtle connotations conveyed by the image. This challenges the model to synthesize information from multiple sources and levels of interpretation to arrive at a holistic understanding.

The rationale for selecting these four tasks is to provide a comprehensive assessment of LVLMs' strengths and weaknesses in interpreting implicit visual meanings. This approach evaluates models across a range of cognitive processes, from basic perception to high-level reasoning and cultural understanding. By structuring the evaluation this way, we gain insights into how well LVLMs mimic human understanding of complex visual stimuli, identify areas for improvement, and guide future research in developing more sophisticated multimodal AI systems capable of nuanced interpretation.

\section{Dataset construction}
\begin{figure*}[t]
    \centering
    \includegraphics[width=0.9\linewidth]{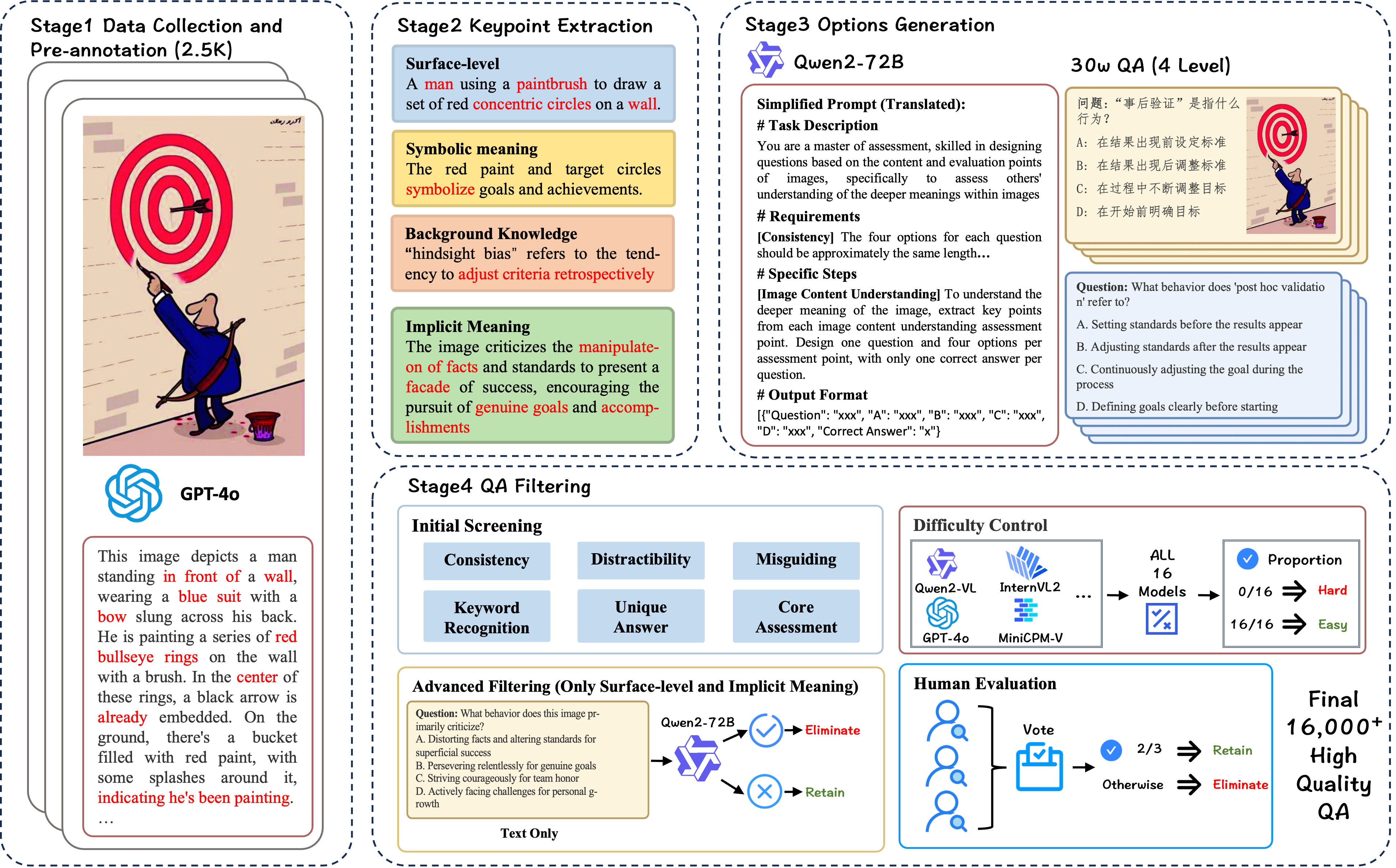}
    
    \caption{InsightVision four-stage construction pipeline. Stage 1 involves data collection and pre-annotation using GPT-4o to generate rich descriptions. Stage 2 conducts keypoint extraction, categorizing information into surface-level content, symbolic meaning, background knowledge, and implicit meaning. Stage 3 utilizes Qwen2-72B for options generation. Finally, Stage 4 applies QA filtering, including consistency checks, difficulty control, and human evaluation, to ensure high-quality, multi-layered annotations.}
    \label{fig:main}
\end{figure*}

Constructing datasets that cover a broad range of knowledge typically requires highly educated annotators, but this approach is time-consuming and costly. To address these challenges, we developed a semi-automatic pipeline for creating the InsightVision dataset, focused on images with implicit meanings. The pipeline includes the following steps (as shown in Figure \ref{fig:main}): 1) Image collection, 2) Data annotation, 3) Keypoint extraction, 4) Question and option generation, and 5) Quality control.
%%% 注意需要添加附录表格。
\subsection{Image collection}
The InsightVision dataset was constructed through a comprehensive web crawling process. We systematically collected approximately 100,000 images from Cartoon Movement\cite{cartoonmovement}, a reputable online platform for editorial cartoons and comics. Each image was accompanied by its associated metadata, including titles, detailed textual descriptions, and relevant keywords. Following the collection phase, we conducted a manual curation process to eliminate duplicates and images lacking implicit meanings. Unlike previous studies, which typically categorize images into a limited set of themes such as humor or satire, we aimed to design a more comprehensive classification system. Therefore, we developed a hierarchical classification system to categorize the curated images based on their primary thematic content. This classification resulted in 13 major categories, including, but not limited to: Arts and Cultural Expression, Economic Development, Social and Cultural Issues, Politics and Power Dynamics, Health and Safety Concerns, and more. These major categories were further subdivided into 41 specific subcategories, providing a granular approach to image classification (Figure \ref{fig:categories}). From these categories, we selected 2,500 images to proceed to the next phase of annotation tasks. We have included a comprehensive list of all categories and subcategories, along with detailed explanations for each, in the Appendix A of this paper.

% This methodology allowed us to construct a diverse and comprehensive dataset that captures a wide range of implicit visual meanings across various socially relevant topics. Our approach combines automated data collection with human-in-the-loop refinement to ensure both scalability and quality in the resulting dataset. %这段看情况删除

\subsection{Data pre-annotation}
To obtain high-quality image annotation data, we implemented a novel approach combining LVLM pre-annotation with human expert verification. This method ensures comprehensive and accurate image understanding, encompassing both explicit visual content and implicit meanings.

\textbf{Pre-annotation model and human annotator selection.} After extensive comparative analysis, we identified GPT-4o as the optimal pre-annotation model. GPT-4o demonstrated superior performance in interpreting nuanced image meanings when provided with textual prompts. To maintain annotation quality, we employed a dual-review process involving two postgraduate-level experts independently verifying each pre-annotation, thus minimizing potential biases and errors.

\textbf{Comprehensive image description generation.} To generate high-quality image understanding data encompassing surface-level content, background knowledge, symbolic meanings, and implicit connotations, we input the crawled images along with their corresponding titles, textual descriptions, and keywords into GPT-4o. Guided by these textual prompts, we instruct GPT-4o to provide a comprehensive description of the image, including: a) Detailed surface-level visual content; b) Implicit meanings and connotations;  c) Requisite background knowledge for understanding these implicit meanings; d) Explanation of symbolic representations and connotations. This approach results in high-quality image-description pairs, each containing a rich, multi-layered interpretation of the visual content. 

\subsection{Keypoint extraction}
After providing a complete description for each image, we extracted key points corresponding to four distinct tasks from these complete descriptions. Each task is exemplified in the Keypoint Extraction box shown in Figure \ref{fig:main}.

% a) Key point of surface-level content include descriptive elements such as "A man using a paintbrush to draw a set of red concentric circles on a wall"; b) Symbolic meaning involves identifying symbolic elements such as "The red paint and target circles symbolize goals and achievements"; c) Background knowledge tasks require the integration of relevant contextual information. For instance, "hindsight bias" refers to the tendency to adjust criteria retrospectively to make outcomes appear successful after they have occurred; d) Implicit meaning focuses on extracting the underlying message, such as "the image criticizes the manipulation of facts and standards to present a facade of success, encouraging the pursuit of genuine goals and accomplishments".

\subsection{Questions and options generation}
After obtaining image annotations, we utilize the annotated keypoints to generate questions and four answer options. Due to the high manual cost, we utilize the complete image descriptions from Section 4.2 and Qwen2-72B to assist in generating questions and options. Qwen2-72B, with 72 billion parameters, is chosen for its capability in natural language generation.

For surface-level understanding, symbolic meaning comprehension, and background knowledge tasks, multiple questions are generated based on keypoints, each with four answer choices. Detailed prompts and examples are provided in Appendix B. For implicit meaning understanding, which primarily evaluates the model’s ability to grasp implicit meanings through reasoning that involves surface-level content, background knowledge, and symbolic interpretation, the answer options tend to be lengthier. As Qwen2-72B's generated questions and answers often diverge excessively, we employ the quality assessment pipeline described in Section 4.5 to enhance the quality of the model-generated questions.

\subsection{Dataset quality}

To ensure dataset quality, we developed a comprehensive set of quality generation criteria and filtering procedures (detailed prompts are provided in the Appendix C).

\textbf{Generation criteria}:

\textbf{1. Consistency.} All options should have roughly the same word count, avoiding obvious length discrepancies. Ensure all options maintain consistency in tone, professionalism, and vocabulary style to prevent the correct answer from being identified through stylistic differences.

\textbf{2. Distractibility.} Wrong options should be designed to be misleading and seemingly reasonable, making them difficult to eliminate by common sense alone. Ensure incorrect options have a certain persuasiveness, rather than being mere assumptions or obvious errors.

\textbf{3. Avoiding image element misguiding.} Ensure that any image elements mentioned in the options match or are similar to the actual content, avoiding easy elimination of incorrect options due to incorrect image details.

\textbf{4. Preventing keyword and pattern recognition.} Avoid obvious keyword matches between the question and options to prevent easy inference.

\textbf{5. Unique correct answer.} Ensure only one correct answer, avoiding ambiguity and ensuring clarity in each option.

\textbf{6. Core assessment.} The design of the question and answer must focus on the key information in the assessment point, which refers to the information related to understanding the deeper meaning.

\textbf{Filtering procedures:}

\textbf{1. Initial filtering.} We employ Qwen2-72B to verify whether generated questions and options fully comply with the six criteria across different understanding levels (surface content, symbolic meaning, background knowledge, and implicit meaning). Questions meeting all criteria are retained; others are regenerated.

\textbf{2. Advanced filtering.} Research suggests that some benchmarks are less reliant on visual input.\cite{tong2024eyes} To ensure true visual dependency and avoid reliance on keyword or pattern recognition, we developed an innovative screening method. Questions are initially input to Qwen2-72B without accompanying images. If the model answers correctly without visual context, the question is discarded and regenerated until it genuinely requires visual input.
%To adhere to preventing keyword and pattern recognition, based on the prior information that shallow content understanding and implicit meaning understanding require visual input for accurate answers, we designed an innovative screening method. 

\textbf{3. Difficulty control.} 
We implemented a model voting system using 16 different models to evaluate question difficulty. The difficulty of each question is determined by the proportion of models that answer it correctly. Questions are categorized based on their correct rate (e.g., 100\% correct rate is classified as easy, 10\% as difficult) and are equally distributed across difficulty levels in the final dataset, excluding the easy level.

\textbf{4. Human evaluation.} Final quality assurance involves a three-person voting system, wherein questions are retained only if all three annotators unanimously agree on their validity and appropriateness. In our study, we recruited a total of nine annotators, who were grouped into teams of three to annotate the same set of questions. The educational backgrounds of the annotators were diverse, comprising two with undergraduate degrees and seven with associate degrees. The questions were broken down into highly specific components, so a high level of academic qualification was not required for annotation.

This methodology ensures a high-quality, visually-dependent dataset with controlled difficulty levels and verified accuracy. Regarding the quality assessment of automatically generated questions, the error rate for pre-annotation using GPT-4 (image pre-annotation) was found to be 2\%, while the final error rate for the generated questions was 5\%. These results suggest that the quality of the automatically generated questions was generally high, demonstrating the effectiveness of the automated process.

\subsection{License and copyright}
% In this dataset, we used the original web links of comic images without infringing their copyrights. Our annotators voluntarily participated in the annotation process and were fairly compensated.
\textbf{Ethics Statement:}
All data samples for this project are sourced from publicly accessible content on social media platforms. To ensure copyright compliance, we use direct links to the original comics to avoid any infringement. Our annotated benchmark will be open-sourced, with links provided for each comic image. We carefully review samples to exclude any content that might be offensive or harmful.
\textit{Additionally, we have obtained permission from the creators to use these public images within our benchmark.}
\newline\textbf{Data Annotation:} Our annotators voluntarily participated in the annotation process and were fairly compensated.

%% file: sec/4_Experiments.tex
\section{Experiments}
Given the impressive performance of LVLMs in tackling image understanding challenges, we evaluated the following LVLMs: InternVL2\cite{chen2024fargpt4vclosinggap}, Qwen2-VL\cite{wang2024qwen2vlenhancingvisionlanguagemodels}, MiniCPM-V-2\_6\cite{yao2024minicpmvgpt4vlevelmllm}, DeepSeek-VL\cite{lu2024deepseekvlrealworldvisionlanguageunderstanding}, LLaVA-OneVision\cite{li2024llavaonevisioneasyvisualtask}, and GPT4o\cite{achiam2023gpt}. These models were selected based on their top-ranking performance in the OpenCompass leaderboard\cite{contributorsopencompass}. Notably, Qwen2-VL-72B\cite{wang2024qwen2vlenhancingvisionlanguagemodels} stands out as the leading open-source LVLMs, while GPT-4o\cite{achiam2023gpt} is widely regarded as one of the excellent closed-source LVLM. Detailed descriptions of these models are provided in the Appendix D.

\subsection{Evaluation}
For evaluating task performance, accuracy was considered the primary metric. A model's answer was deemed correct if it matched the ground truth. Accuracy was computed as the ratio of the number of correct answers ($N_{r}$) to the total number of questions ($N$), i.e.,$N_{r}/N$].

Our task prompts were determined based on each image and task type (referring to the four tasks), followed by choice options: A, B, C, D. The specific parameter settings, including temperature and top-k values, used for each model in the experiments are detailed in the Appendix E. Furthermore, to assess human performance on these tasks, we randomly selected 100 questions from each task in the dataset and had human evaluators provide answers. This allowed us to benchmark human participants' performance against our models, providing a comprehensive comparison of human and machine capabilities on these specific tasks. Detailed experimental results are shown in Table \ref{tab:benchmark}.

% \subsection{Main Results}
% \usepackage{tabularray}
\begin{table*}
\centering
\begin{tblr}{
  width = \linewidth,
  colspec = {Q[300]Q[94]Q[133]Q[92]Q[121]Q[63]Q[140]},
  column{even} = {c},
  column{3} = {c},
  column{5} = {c},
  column{7} = {c},
  hline{1,17} = {-}{0.08em},
  hline{2} = {-}{0.05em},
  hline{14} = {1}{l},
  hline{14} = {2-6}{},
  hline{14} = {7}{r},
  hline{15} = {1}{l},
  hline{15} = {2-6}{},
  hline{15} = {7}{r},
  hline{16} = {1}{l},
  hline{16} = {2-6}{},
  hline{16} = {7}{r},
  % hline{17} = {1}{l},
  % hline{17} = {2-6}{},
  % hline{17} = {7}{r},
  % hline{16-17} = {1}{l},
  % hline{16-17} = {2-6}{},
  % hline{16-17} = {7}{r},
  %hline{1,19} = {-}{0.08em},
  %hline{2} = {-}{0.05em},
  %hline{17-18} = {1}{l},
  %hline{17-18} = {2-6}{},
  %hline{17-18} = {7}{r},
}
\textbf{Model}                 & \textbf{\# Params} & \textbf{Surface} & \textbf{Symbolic} & \textbf{Background} & \textbf{Mean} & \textbf{Implicit} \\
InternVL2-Llama3-76B\cite{chen2024fargpt4vclosinggap}           & 76B                & 74.7                   & 71.1              & 75.4                 & 73.7          & 53.8                      \\
Qwen2-VL-72B-Instruct\cite{wang2024qwen2vlenhancingvisionlanguagemodels}            & 72B                & 79.3                   & \textbf{82.6}     & \textbf{81.6}        & \textbf{81.2} & \textbf{60.1}             \\
InternVL2-40B\cite{chen2024fargpt4vclosinggap}                  & 40B                & 79.5                   & 79.8              & 80.7                 & 80.0          & 58.7                      \\
InternVL1.5-26B\cite{chen2024fargpt4vclosinggap}                & 26B                & 74.1                   & 70.5              & 74.4                 & 73.0          & 54.7                      \\
InternVL2-26B\cite{chen2024fargpt4vclosinggap}                  & 26B                & 75.2                   & 71.8              & 73.9                 & 73.6          & 50.7                      \\
InternVL2-8B\cite{chen2024fargpt4vclosinggap}                   & 8B                 & 70.7                   & 73.6              & 73.7                 & 72.7          & 46.5                      \\
MiniCPM-V-2\_6\cite{yao2024minicpmvgpt4vlevelmllm}                   & 8B                 & 74.0                   & 74.1              & 79.2                 & 75.8          & 50.0                      \\
Qwen2-VL-7B-Instruct\cite{wang2024qwen2vlenhancingvisionlanguagemodels}             & 7B                 & 75.1                   & 81.1              & 79.3                 & 78.5          & 51.7                      \\
llava-onevision-qwen2-7b\cite{li2024llavaonevisioneasyvisualtask}  & 7B                 & 74.2                   & 72.9              & 76.2                 & 74.4          & 50.0                      \\
v2\_deepseek-vl-7b-chat\cite{lu2024deepseekvlrealworldvisionlanguageunderstanding}        & 7B                 & 58.8                   & 57.3              & 65.6                 & 60.6          & 38.1                      \\
% InternVL2-4B\cite{chen2024fargpt4vclosinggap}                   & 4B                 & 63.3                   & 68.9              & 69.8                 & 67.3          & 41.3                      \\
% InternVL2-2B\cite{chen2024fargpt4vclosinggap}                   & 2B                 & 58.2                   & 56.0              & 64.4                 & 59.5          & 37.6                      \\
Qwen2-VL-2B-Instruct\cite{wang2024qwen2vlenhancingvisionlanguagemodels}           & 2B                 & 70.3                   & 73.8              & 74.5                 & 72.9          & 45.2                      \\
llava-onevision-qwen2-0.5b\cite{li2024llavaonevisioneasyvisualtask} & 0.5B                 & 44.4                   & 45.0              & 33.3                 & 40.9          & 23.2                      \\
% InternVL2-1B\cite{chen2024fargpt4vclosinggap}                   & 1B                 & 49.9                   & 56.5              & 57.9                 & 54.8          & 40.3                      \\
GPT4o                          & -                  & \textbf{82.0}          & 80.8              & 79.8                 & 80.9          & 59.3                      \\
% Ours                          & 7B                  & 78.4          & \textbf{84.6}              & \textbf{83.9}                 & \textbf{82.3}          & \textbf{62.5}                      \\

Human                          & -                  & 98.0                   & 88.0              & 86.0                 & 90.7          & 74.0                      
\end{tblr}
\caption{The benchmark includes the average accuracy (in percentages (\%)) on four tasks. Surface, Symbolic, Background, and Implicit represent Surface-level Content Understanding Task, Symbolic Meaning Interpretation Task, Background Knowledge Comprehension Task, and Implicit Meaning Comprehension Task, respectively. The Mean represents the average accuracy of the first three tasks.}
\label{tab:benchmark}
\end{table*}

\subsection{Main results}
\textbf{Surface-level content understanding.} Among the open-source models, Qwen2-VL-72B-Instruct and InternVL2-40B performed best on the surface-level content understanding task, with accuracies of 79.3\% and 79.5\%, respectively, close to GPT-4o (82.0\%). Performance generally correlated with model size, ranging from 44.4\% for the 0.5B llava-onevision-qwen2 to 79.3\% for the 72B Qwen2-VL. However, all models showed a substantial gap compared to human performance (98.0\%), highlighting room for improvement.

\textbf{Symbolic meaning interpretation.} Qwen2-VL-72B-Instruct performed optimally, achieving an accuracy of 82.6\%, slightly surpassing GPT4o's 80.8\%. Smaller models like llava-onevision-qwen2-0.5b-ov-hf achieved only 45.0\%, suggesting that model scale significantly impacts symbolic understanding capabilities. Most models' performance on this task was similar to the surface-level content understanding task, indicating comparable difficulty levels for symbolic meaning interpretation and surface-level content understanding.

\textbf{Background knowledge comprehension.} InternVL2-40B and Qwen2-VL-72B-Instruct exhibited the best performance, with accuracies of 80.7\% and 81.6\%, respectively. The relatively small gap compared to human performance (86.0\%) indicates that models have made significant progress in background understanding. %Even smaller models, such as InternVL2-1B, achieved 58\% accuracy, suggesting that background knowledge comprehension is not entirely dependent on model scale.

\textbf{Implicit meaning comprehension.} All models performed significantly worse on the implicit meaning comprehension task compared to the other tasks. The best performance was achieved by Qwen2-VL-72B-Instruct at 60.1\%, comparable to GPT4o (59.3\%). Smaller models like llava-onevision-qwen2-0.5b-ov-hf achieved only 23\%, revealing a substantial gap compared to human performance (74.0\%). This task appears to be the most challenging for current LVLMs.

\section{Analysis}

\subsection{How do the models perform across different categories of visual perception?}
\begin{figure*}[t]
    \centering
    \includegraphics[width=0.65\linewidth,height=0.45\linewidth]{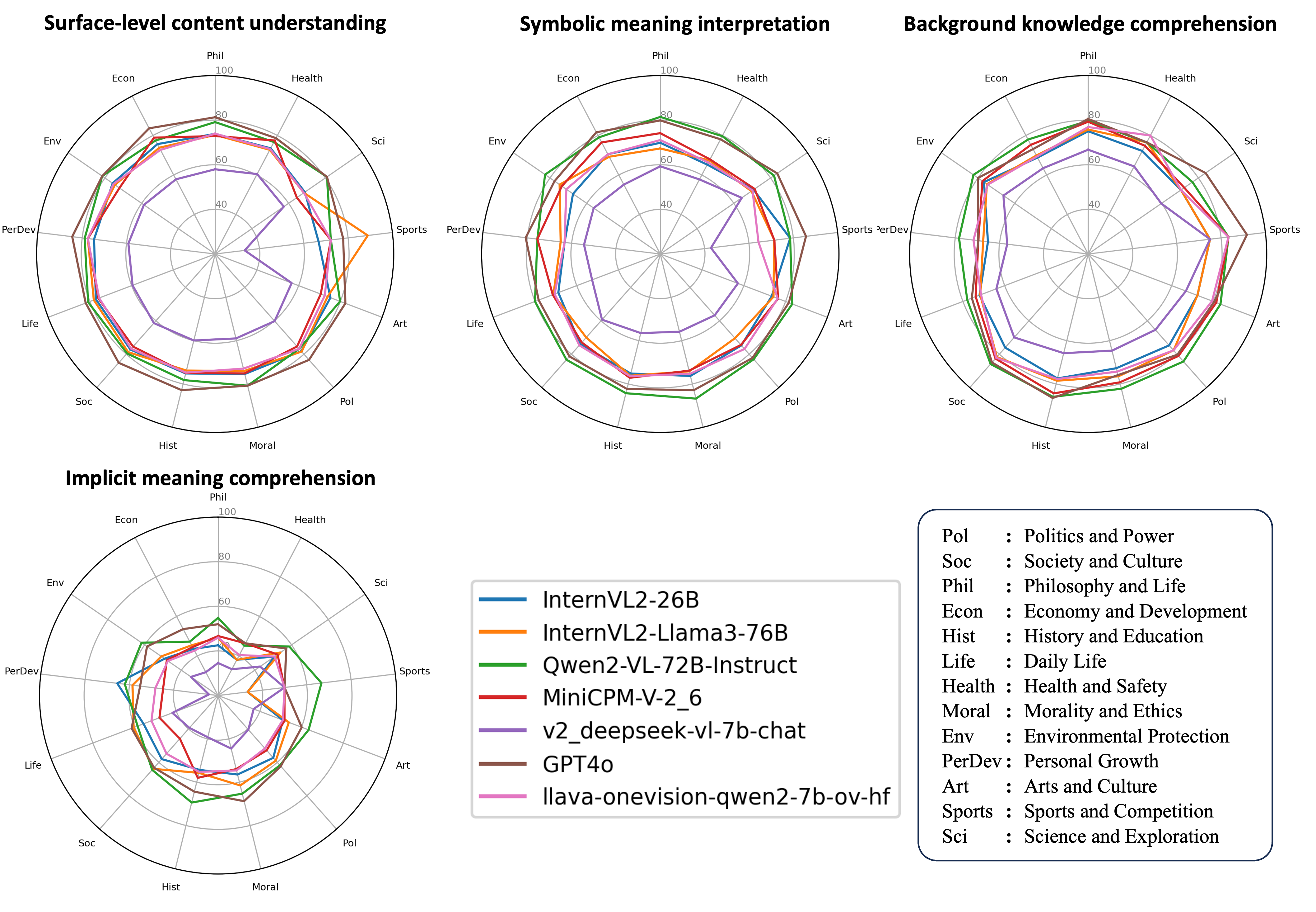}
    
    \caption{The radar charts illustrate the performance of various representative models in interpreting images across different categories within our four tasks.}
    \label{fig:radar}
\end{figure*}
% \begin{figure*}[t]
%     \centering
%     \includegraphics[width=1\linewidth]{file/radar_3.png}
    
%     \caption{The radar charts illustrate the performance of various representative models in interpreting images across different categories within our four tasks.}
%     \label{fig:datasetoverview}
% \end{figure*}
% How does the model parameter scale affect deep semantic understanding?}
% Due to the scaling law, the number of parameters typically has a positive impact on model performance. In this context, we also discuss the relationship between model parameter scale and deep semantic understanding. We examined two pairs of models: the InternVL2 series and the Qwen2-VL sequence. Models within the same series have a consistent architecture and training process, differing only in parameter scale. Figure \ref{fig:bar} provides the accuracy of the models from the two series on four tasks. It can be observed that as the number of model parameters increases, there is a consistent improvement in performance across all four tasks. Therefore, an increase in the number of parameters has a positive impact on the model's deep semantic understanding capability.
Figure \ref{fig:radar} illustrates model performance across four key tasks: surface-level content understanding, symbolic meaning interpretation, background knowledge comprehension, and implicit meaning comprehension, spanning various categories. Accuracy varies significantly across categories. In simpler categories like history and environment, models achieve higher accuracy by effectively capturing direct information. However, performance drops in categories involving deep cultural symbols or metaphors, such as philosophy and personal growth, highlighting current models' limitations in handling complex semantics and cultural nuances.

Larger models (40B+) consistently outperform smaller ones, especially on complex tasks. For simpler tasks like surface-level content, all models perform well, though larger models still have an edge. As task complexity increases, performance gaps widen, with top models significantly surpassing smaller ones but still facing challenges. The Qwen2-VL and InternVL2 series excel in symbolic meaning and background knowledge but show varying stability in implicit meaning comprehension, highlighting ongoing challenges in complex semantic interpretation. These results suggest that while scaling improves performance, implicit meaning comprehension requires further architectural or training optimizations for substantial progress.

\subsection{Can Image Descriptions Help the Model Understand Implicit Meaning?}
\begin{figure}[t]
  \centering
  \includegraphics[width=1\linewidth, height=0.7 \linewidth]{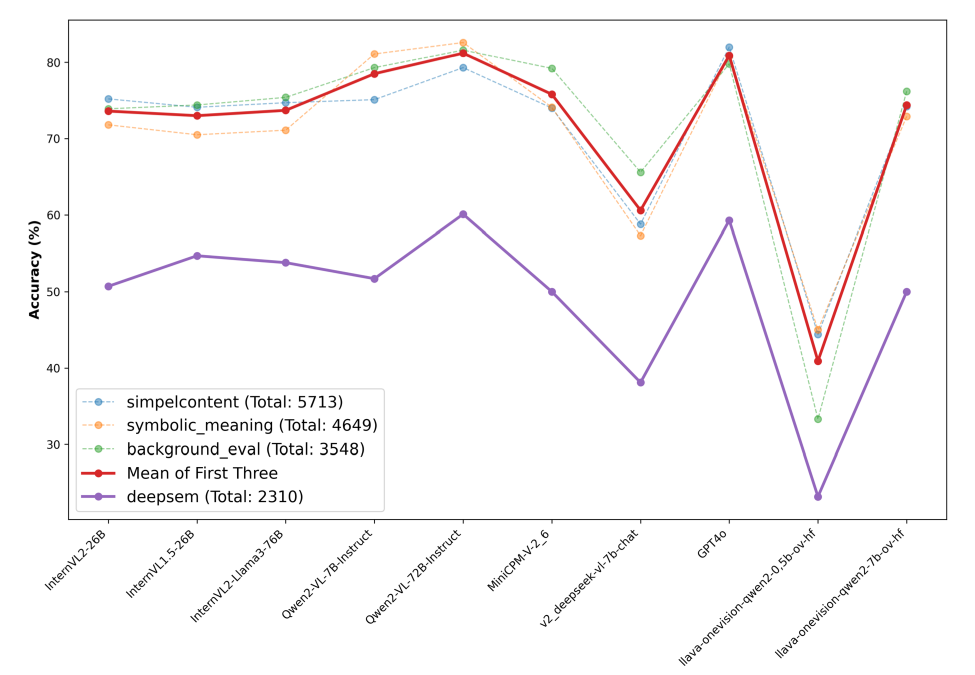}
   \caption{Relationship between implicit meaning comprehension and other tasks.}
   \label{fig:6.2}
\end{figure}
%Our analysis shows that providing detailed image descriptions enhances LVLMs' ability to understand deeper semantics. Figure \ref{fig:6.2} illustrates that performance in implicit meaning comprehension is closely linked to the performance in the first three tasks: surface-level content, symbolic meaning, and background knowledge comprehension. The results demonstrate that models provided with additional descriptive information consistently achieve higher accuracy compared to those without, confirming that added context facilitates deeper understanding(see Appendix F for detailed results). However, while descriptions aid models, they still fall short of human-level understanding, particularly in implicit meaning comprehension tasks. This suggests that while auxiliary descriptions help, further advancements are needed for true human-like comprehension.
We believe that, like humans, models need to combine surface-level content, symbolic meaning, and background knowledge to understand implicit meanings. Figure \ref{fig:6.2} shows that performance in implicit meaning comprehension is closely related to the first three tasks. To further validate this, we added key information from these tasks to the reasoning prompts. Experimental results (see Appendix F) show significant improvement, with the optimal model's accuracy surpassing human performance. We reasonably assume that adding this information enables the model to capture most of the foundational content and background knowledge required for implicit meaning comprehension.  However, despite these benefits, there remains room for improvement, suggesting that capturing key information alone is insufficient for fully understanding implicit meanings. To achieve human-like comprehension, models need not only the ability to capture key information but also the reasoning ability to process it effectively.

\subsection{How Does Model Parameter Scale Affect Implicit Meaning Comprehension?}
\begin{figure}[t]
  \centering
  \includegraphics[width=1.0 \linewidth]{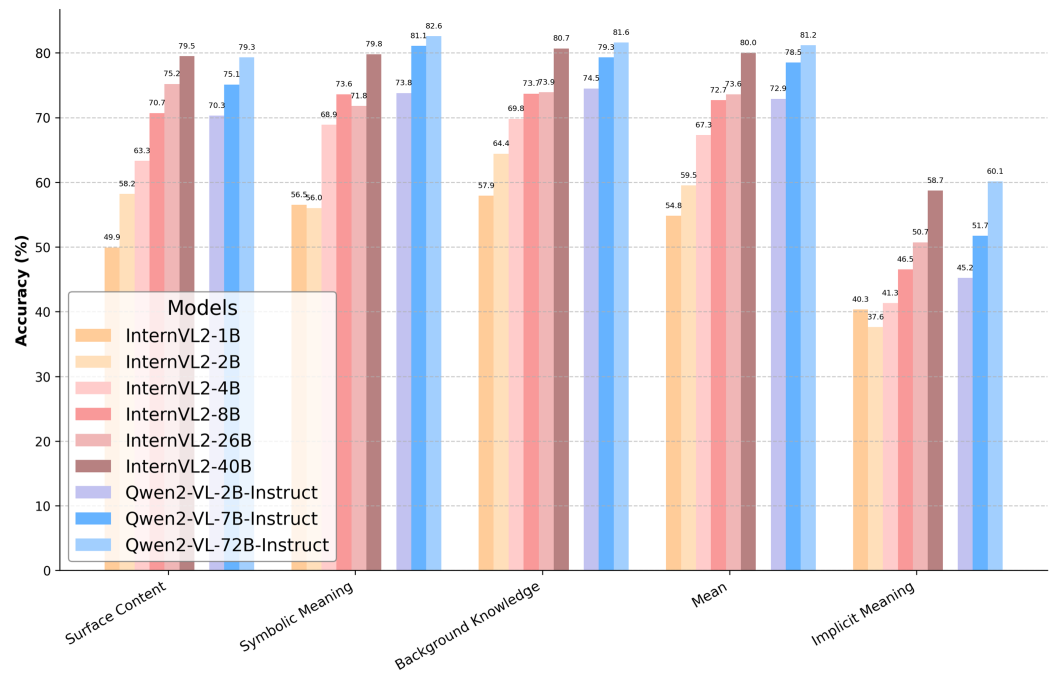}
   \caption{Comparison of accuracy across tasks for InternVL2 models (1B to 40B) and Qwen2-VL-Instruct models (2B to 72B)}
   \label{fig:bar}
\end{figure}

According to scaling laws, increasing model parameters generally improves performance. To evaluate this relationship, we selected models of different scales from two distinct series, InternVL2 and Qwen2-VL. Within each series, the models share the same architecture but differ in scale. InternVL2 and Qwen2-VL, which share architecture but vary in size. Figure \ref{fig:bar} shows that larger models perform better across all four tasks, with models in the 40B-72B range balancing performance and computational cost. However, deeper semantic tasks may need further architectural optimizations, indicating that enhancing deep semantic comprehension requires more than scaling—it also needs specialized strategies.

%由于漫画通常具有高度浓缩的视觉信息，忽略细节往往会导致理解与作者意图之间的巨大差异。同时，漫画中的符号和象征意义常常依赖特定的文化和历史背景，存在一定的理解门槛。此外，一些漫画的含义过于抽象，要求更高的推理和解读能力，降低了人工测试的准确率。
%table need adjustment

%% file: sec/5_Conclusion.tex
\section{Conclusion}
\label{sec:Conclusion}
We introduce InsightVision, a comprehensive, multi-level Chinese-based benchmark designed to evaluate the understanding of implicit visual semantics in LVLMs. The benchmark comprises over 2,500 carefully curated images, each paired with questions that assess four levels of comprehension: surface-level content understanding, symbolic meaning interpretation, background knowledge comprehension, and implicit meaning comprehension. Our evaluations demonstrate a considerable gap between current LVLMs and human performance, particularly in understanding implicit meanings. We suggest that enhancing model parameters or integrating detailed image descriptions during reasoning may help improve the model's ability to capture and interpret deeper semantic content. This work underscores the need for more advanced multimodal models capable of nuanced visual semantic understanding. We hope InsightVision will serve as a valuable resource for advancing research aimed at bridging the gap between perceptual recognition and cognitive understanding of visual content.

\section*{Limitations}
The InsightVision dataset currently focuses on comic images, which effectively convey implicit meanings but lack visual diversity. Future expansions will include other media, such as photography and video, to enhance diversity and applicability. Additionally, the dataset is based on Chinese cultural contexts, which may limit generalizability; broader cultural inclusion is planned. Lastly, despite using GPT-4o and human review for annotation, biases and errors may still exist, and improvements to the generation pipeline are needed to address these issues.
% The InsightVision dataset currently focuses exclusively on comic images, selected for their ability to convey rich, implicit meanings. While comics are effective for evaluating complex visual semantics, this choice may limit the diversity of visual styles represented. Future work will expand the dataset to include other forms of visual media, such as photography and video, to more fully capture real-world visual scenarios.

% Furthermore, InsightVision is predominantly framed within Chinese cultural contexts, which could restrict its applicability to models trained on or intended for different cultural backgrounds. Expanding the dataset to include a wider array of cultural perspectives is an important avenue for future research.

% Finally, while pre-annotation models like GPT-4o are effective in generating image annotations, they may introduce biases. Despite implementing a dual-review process with human verification, the potential for annotation errors and biases remains. The dataset’s generation pipeline can be further refined to mitigate these issues, ensuring a more robust and unbiased benchmark.

\section*{Acknowledgements}
The authors wish to thank the anonymous reviewers for their helpful comments. This work was supported by Ant Group Research Intern Program.

%% file: sec/appendix.tex
\appendix

\clearpage
\setcounter{page}{1}
\maketitlesupplementary

\section{Categories Definition}
The hierarchical classification system mentioned in Section 4.1 is detailed as follows. We first instruct GPT-4o to output the potential categories and corresponding subcategories for the comic images. Then, we provide the collected 100,000 comic images to GPT-4o to classify them into the newly formed categories. A significant portion of the images will not find a corresponding category. We then instruct GPT-4o to complete the classification based on the remaining images, and reclassify the comic images. This process is repeated until all images are classified. The final result is shown in Table \ref{tab:category}, which includes 13 categories, 41 subcategories, and their corresponding specific definitions.

\begin{table*}
\centering
\resizebox{\textwidth}{!}{
\begin{tblr}
{
  row{1} = {c},
  cell{2}{1} = {r=6}{c},
  cell{2}{2} = {c},
  cell{3}{2} = {c},
  cell{4}{2} = {c},
  cell{5}{2} = {c},
  cell{6}{2} = {c},
  cell{7}{2} = {c},
  cell{8}{1} = {r=4}{c},
  cell{8}{2} = {c},
  cell{9}{2} = {c},
  cell{10}{2} = {c},
  cell{11}{2} = {c},
  cell{12}{1} = {r=2}{c},
  cell{12}{2} = {c},
  cell{13}{2} = {c},
  cell{14}{1} = {r=2}{c},
  cell{14}{2} = {c},
  cell{15}{2} = {c},
  cell{16}{1} = {r=3}{c},
  cell{16}{2} = {c},
  cell{17}{2} = {c},
  cell{18}{2} = {c},
  cell{19}{1} = {r=3}{c},
  cell{19}{2} = {c},
  cell{20}{2} = {c},
  cell{21}{2} = {c},
  cell{22}{1} = {r=3}{c},
  cell{22}{2} = {c},
  cell{23}{2} = {c},
  cell{24}{2} = {c},
  cell{25}{1} = {r=4}{c},
  cell{25}{2} = {c},
  cell{26}{2} = {c},
  cell{27}{2} = {c},
  cell{28}{2} = {c},
  cell{29}{1} = {r=4}{c},
  cell{29}{2} = {c},
  cell{30}{2} = {c},
  cell{31}{2} = {c},
  cell{32}{2} = {c},
  cell{33}{1} = {r=2}{c},
  cell{33}{2} = {c},
  cell{34}{2} = {c},
  cell{35}{1} = {r=3}{c},
  cell{35}{2} = {c},
  cell{36}{2} = {c},
  cell{37}{2} = {c},
  cell{38}{1} = {r=3}{c},
  cell{38}{2} = {c},
  cell{39}{2} = {c},
  cell{40}{2} = {c},
  cell{41}{1} = {r=2}{c},
  cell{41}{2} = {c},
  cell{42}{2} = {c},
  vline{2-3} = {1-42}{0.05em},
  vline{3} = {3-7,9-11,13,15,17-18,20-21,23-24,26-28,30-32,34,36-37,39-40,42}{0.05em},
  hline{1,43} = {-}{0.08em},
  hline{2,8,12,14,16,19,22,25,29,33,35,38,41} = {-}{0.05em},
  hline{3-7,9-11,13,15,17-18,20-21,23-24,26-28,30-32,34,36-37,39-40,42} = {2-3}{dashed},
}
\textbf{Category} & \textbf{Subcategory} & \textbf{Definition}\\
Politics and Power & Political Games & Policy disputes, party struggles, concentration of power.\\
 & Political Corruption & Corruption, abuse of power, electoral fraud.\\
 & Political Figures & Behaviors of leaders, public images, personal scandals.\\
 & National Situation & {International relations, national divisions, territorial disputes, independence movements, ethnic \\conflicts.}\\
 & National Symbols and Dignity & Actions like damaging the national flag, emblem, or offending national symbols.\\
 & Freedom of Speech and Media & Issues related to freedom of speech, press freedom, censorship, and information control.\\
Society and Culture & Social Phenomena & {Racism and sexism, consumerism, celebrity scandals, cults, extremist religious groups, celebrity \\worship, superheroes as cultural symbols, conflicts in sports competitions.}\\
 & Cultural Phenomena & Modern lifestyles, technological dependency, pop culture, and media commentary.\\
 & Social Inequality & Wealth gap, labor rights, social stratification.\\
 & Protection of Minors & Issues like harmful animations, violence, and soft-pornography involving minors.\\
Economy and Development & Economic Issues & Economic crises, wealth inequality, impacts of globalization.\\
 & Technological Development & {Privacy concerns, tech monopolies, ethics in technology, future technologies, cybersecurity, tech \\and space exploration, innovation.}\\
History and Education & Historical Events & Wars, revolutions, significant historical events.\\
 & Educational Issues & Education equity, academic misconduct, reforms, and pressures.\\
Daily Life & Family Relationships & Family conflicts, generational differences, marriage issues, family humor, and reconciliation.\\
 & Work Environment & Workplace issues, corporate culture, job stress.\\
 & Leisure and Celebrations & Festivals, celebrations, summer leisure, and reading.\\
Health and Safety & Public Health & Pandemics, healthcare systems, vaccinations.\\
 & Food Safety & GMOs, food additives.\\
 & Mental Health & Issues like suicide, self-harm, depression, and anxiety.\\
Morality and Ethics & Social Morality & Hypocrisy, greed, selfishness.\\
 & Sex and Morality & Sexual behaviors, innuendos, gender discrimination, sex scandals.\\
 & Tech Ethics & Artificial intelligence, genetic editing, comparison between science and pseudoscience.\\
Environmental Protection & Environmental Pollution & Air, water, and soil pollution.\\
 & Ecological Damage & Deforestation, ocean pollution, loss of biodiversity.\\
 & Climate Change & Global warming, extreme weather events.\\
 & Sustainable Development & Resource management, green technologies, environmental policies.\\
Arts and Culture & Artistic Creation and Expression & Artistic techniques, symbolic meanings, art and technology, visual language of art and symbols.\\
 & Art and Philosophy & Surrealism, philosophy and art, existentialism in art.\\
 & Art and Culture & Modern art, geometric abstraction, cultural commemorations, semiotics in art.\\
 & Art and Entertainment & Music, films, games as forms of celebration and artistic entertainment.\\
Sports and Competition & Sports Events & Sports competitions, achievements, and glory.\\
 & Sports and Culture & Team spirit, artistic and entertaining aspects of sports.\\
Science and Exploration & Scientific Research & Unknowns of scientific exploration, satire on pseudoscience, cognition and science.\\
 & Exploration and Mysteries & {Adventures, harmony of nature and urban civilizations, space exploration, and international coop-\\eration.}\\
 & Philosophy and Science & Philosophy of time, cosmology, intersections of science and philosophy.\\
Philosophy and Life~ & Existence and Reflection & Wisdom and loneliness, symbols of creativity, existentialism.\\
 & Psychology and Emotion & Emotions and remembrance, perseverance in adversity, happiness and contentment.\\
 & Time and Life & Time and life, philosophy of time, time management and psychological adaptation.\\
Personal Grow & Personal Growth & Challenges, effort and success, self-care.\\
 & Creativity and Inspiration & Creative thinking, capturing inspiration, art and creativity, overcoming challenges with wisdom.
\end{tblr}
}
\caption{The names and detailed definitions of the categories and subcategories in InsightVision}
\label{tab:category}
\end{table*}

\section{Prompt}
The \{Comprehensive image description\} in the prompt refers to the process in Section 4.2 where we instruct GPT-4o to provide a comprehensive description of the image, including:a) Detailed surface-level visual content;b) Implicit meanings and connotations;c) Requisite background knowledge for understanding these implicit meanings; d) Explanation of symbolic representations and connotations
\subsection{Implicit meaning summarization}
\label{subsection:B.1}

\begin{figure}[t]
  \centering
  \includegraphics[width=1.0 \linewidth]{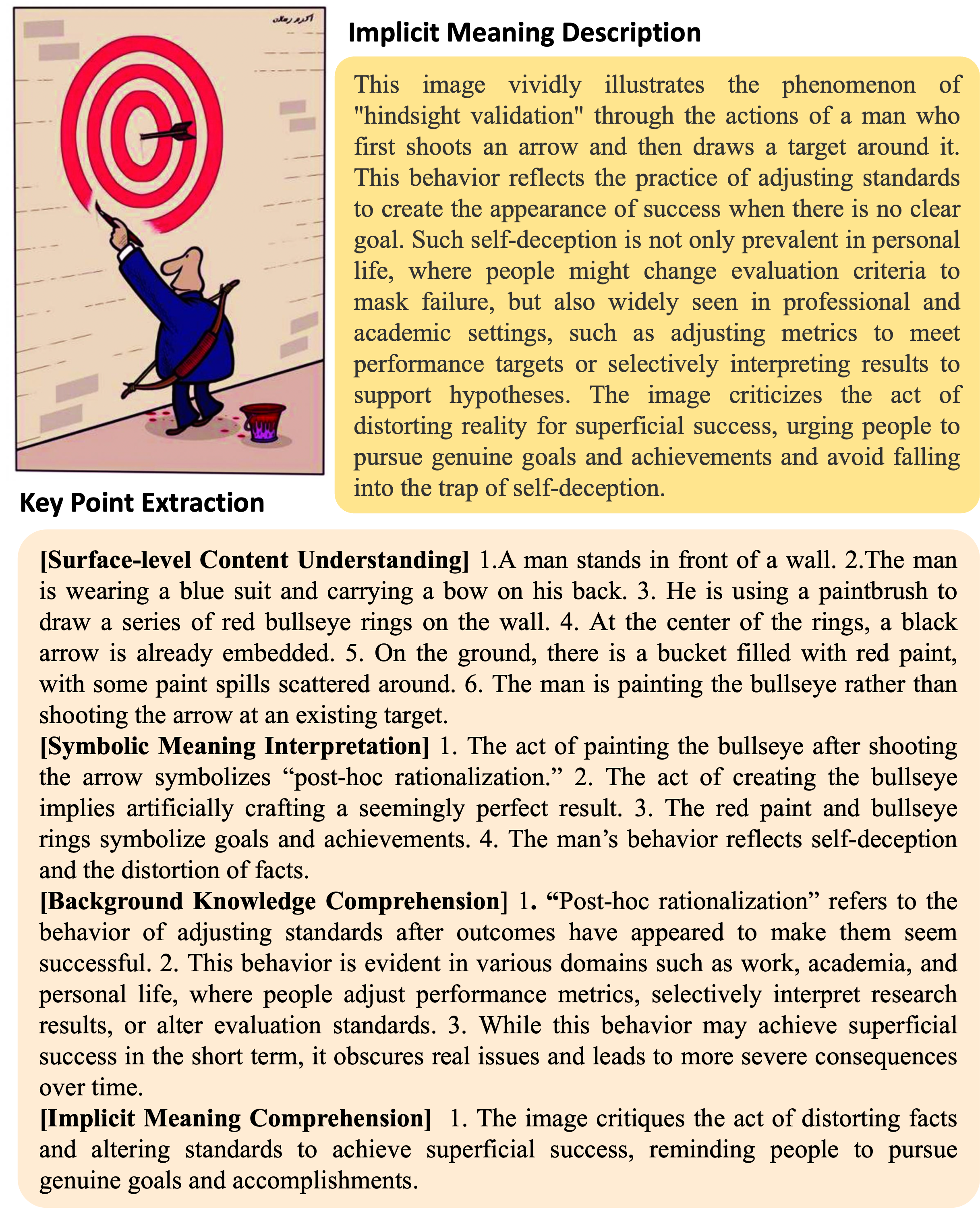}
   \caption{Example of Implicit meaning description and Key point extraction}
   \label{fig:example1}
\end{figure}
To enable the LLM to summarize the implicit meaning of an image, we use the following prompt (originally in Chinese, but shown here in English).The example of output is illustrated in Figure \ref{fig:example1}.
\begin{tcolorbox}[colback=gray!5,colframe=gray!80, breakable,title=Prompt]
\textbf{\# Task Description}
\newline You are a master of understanding the implicit meaning of images and need to accurately summarize the profound implications of the input image.
\newline \textbf{\# Specific Requirements}
\newline Based on the elements and related details in the image, identify and accurately summarize the deep meaning. Output only the summary without any additional symbols.
\newline \textbf{\# Image Content }(Although you cannot see the image, I will describe it in the following text)
\newline \{Comprehensive image description\}
\end{tcolorbox}

\subsection{Key point extraction}
\label{subsection:B.2}
Based on the implicit meaning concluded in \ref{subsection:B.1} (referred to as \{Implicit meaning\} in the prompt) , we extract key point by the following prompt(originally in Chinese, but shown here in English).The model's output example is illustrated in the lower part of Figure \ref{fig:example1}.

\begin{tcolorbox}[colback=gray!5,colframe=gray!80, breakable,title=Prompt]
\textbf{\# Task Description}
\newline You are a master of logical reasoning and can infer the deep meaning of an image based on its content. Now, given the image content and the deep meaning, analyze step by step to identify the key elements needed to infer the deep meaning from the image content.
\newline\textbf{\# Specific Steps}
\newline  Follow these steps for the analysis:
\newline [Surface-level Content Understanding] Understand the image content that is necessary to grasp the deep meaning, mainly including the facts depicted in the image.
\newline [Symbolic Meaning Interpretation] Understand the symbolic, implicit, metaphorical, suggestive, or potential meanings that are abstract and related to the deep meaning of the image.
\newline [Background Knowledge Comprehension] Identify the specific historical knowledge or relevant common sense required to understand the deep meaning of the image, without abstract concepts.
\newline [Implicit Meaning Comprehension] Summarize the deep meaning of the image in a short phrase or sentence.
\newline \textbf{\# Output Format}
\newline[Surface-level Content Understanding] 1.xxx; 2.xxx; ...
\newline[Symbolic Meaning Interpretation] 1.xxx; 2.xxx; ...
\newline[Background Knowledge Comprehension] 1.xxx; 2.xxx; ...
\newline[Implicit Meaning Comprehension] xxx
\newline \textbf{ \# Image Content}(Although you cannot see the image, I will describe it in the following text)
\newline \{Comprehensive image description\}
\newline \textbf{\# Deep Meaning}
\newline \{Implicit meaning\}

\end{tcolorbox}

\subsection{QA Generation}
\label{subsection:B.3}
\begin{figure*}[t]
  \centering
  \includegraphics[width=0.9 \linewidth]{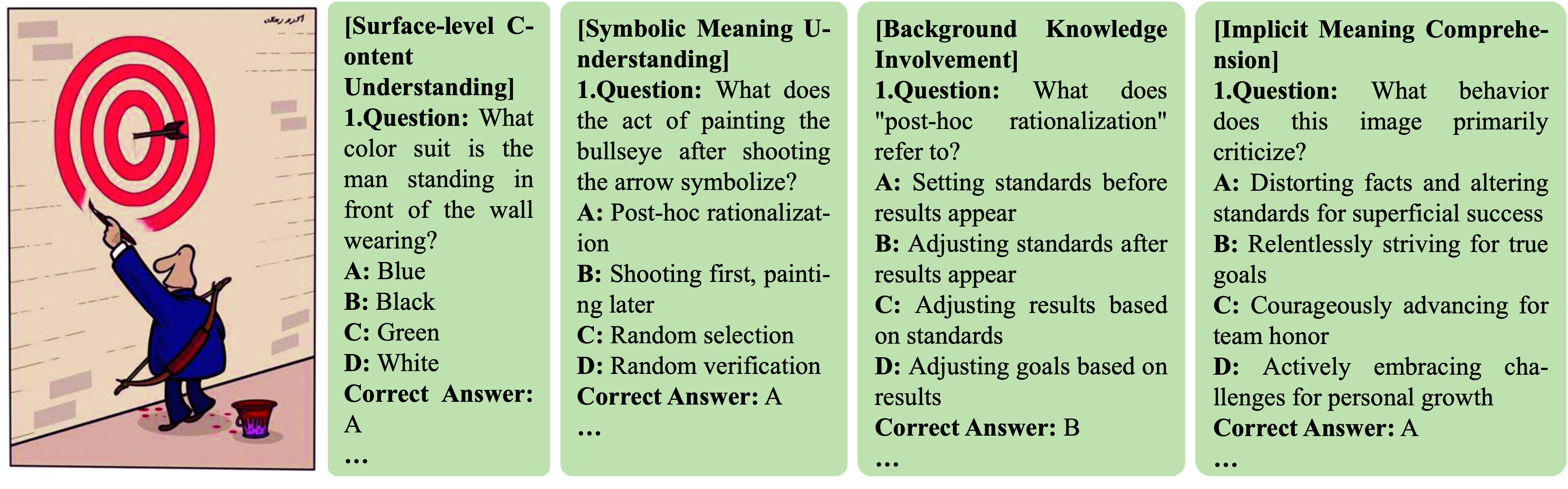}
   \caption{Examples of QA Generation}
   \label{fig:example2}
\end{figure*}

To generate high-quality QA, we explicitly inform the LLM of six requirements in the prompt and specify the analysis steps and output format. The complete prompt is shown below(originally in Chinese, but shown here in English), and the corresponding output is illustrated in Figure \ref{fig:example2}.
The \{Key points\} in the prompt refers to the output of section \ref{subsection:B.2} with examples shown in the lower part of Figure \ref{fig:example1}.The Implicit Meaning in the prompt refers to the output from B.1, as shown in the upper part of Figure \ref{fig:example1}.

\begin{tcolorbox}[colback=gray!5,colframe=gray!80, breakable,title=Prompt]
\textbf{\# Task Description}
\newline You are an evaluation master, skilled in designing questions based on image content and assessment points, specifically to test others' understanding of the deeper meaning of images.
\newline\textbf{\# Requirements }(All questions and options must meet the following requirements)
\newline These requirements ensure that the designed questions and options can effectively assess whether others understand the key points without being influenced by formal differences.
\newline[Consistency] Ensure that the four options under the same question are approximately the same length to avoid obvious differences in length. Maintain a consistent tone and style across the four options, ensuring similar word choices to prevent identifying the correct option through stylistic differences.
\newline[Distractibility] Wrong options should be confusing and seemingly reasonable, making them not easily ruled out by common sense. Ensure that wrong options are somewhat persuasive, not just hypothetical or obviously incorrect.
\newline[Avoiding Image Element Misguiding] Ensure that the image elements mentioned in the four options match or are similar to the actual content, to avoid easily ruling out wrong options due to incorrect image details.
\newline[Preventing Keyword and Pattern Recognition] Avoid obvious keyword matches between questions and options. Ensure there is no direct verbal association between the question and the correct option to prevent easy inference.
\newline[Unique Correct Option] Among the four options, ensure that only one is the correct option. Avoid ambiguity or vagueness, allowing each option to have a clear judgment.
\newline[Core Assessment] The design of questions and options must be based on the "key points."
\newline\textbf{\# Specific Steps}
\newline Analyze according to the following steps:
\newline[Surface-level Content Understanding] To understand the deeper meaning of the image, extract key points from each image content understanding assessment point. Design a question and four options for each assessment point, with only one correct option per question.
\newline[Symbolic Meaning Interpretation] To understand the deeper meaning of the image, extract key points from each symbolic meaning understanding assessment point. Design a question and four options for each assessment point, with only one correct option per question.
\newline[Background Knowledge Comprehension] To understand the deeper meaning of the image, extract key points from each background knowledge involvement assessment point. Design a question and four options for each assessment point, with only one correct option per question.
\newline[Implicit Meaning Comprehension] Design a question and four options based on the deep meaning assessment point of the image, with only one correct option.
\newline\textbf{\# Output Format} (Please strictly follow the format below)
\newline\{ "Surface-level Content Understanding":
\newline[\{"Question":"xxx","A":"xxx","B":"xxx","C":"xxx",\newline"D":"xxx","Correct Option":"x"\}, \{...\}], 
\newline"Symbolic Meaning Interpretation":
\newline[\{"Question":"xxx","A":"xxx","B":"xxx","C":"xxx",\newline"D":"xxx","Correct Option":"x"\}, \{...\}], 
\newline"Background Knowledge Comprehension":
\newline[\{"Question":"xxx","A":"xxx","B":"xxx","C":"xxx",\newline"D":"xxx","Correct Option":"x"\}, \{...\}], 
\newline"Implicit Meaning Comprehension":
\newline[\{"Question":"xxx","A":"xxx","B":"xxx","C":"xxx",\newline"D":"xxx","Correct Option":"x"\}] \}
\newline \textbf{\#Image Content} (although you cannot see the image, I will describe the image with the following text)
\newline \{Comprehensive image description\}
\newline \textbf{\# Assessment Points List}
\newline \{Key points\}
\newline \textbf{\# Deep Meaning}
\newline \{Implicit meaning\}
\end{tcolorbox}

\section{QA Filtering}
\subsection{Initial screening}
\label{subsection:C.1}
\begin{figure}[t]
  \centering
  \includegraphics[width=1.0 \linewidth]{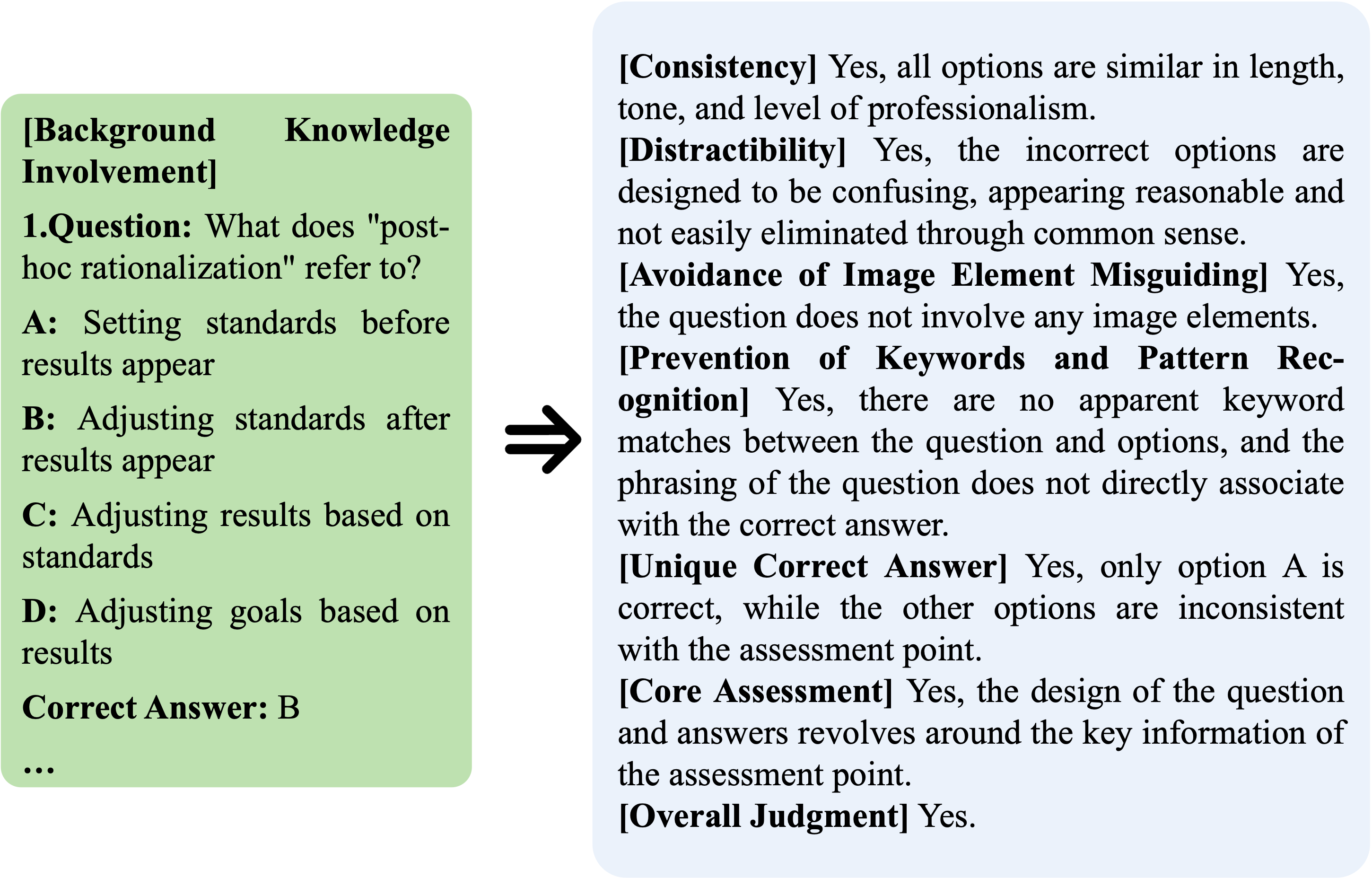}
   \caption{Example of Initial screening}
   \label{fig:example3}
\end{figure}
For the QA generated according to the specified requirements in \ref{subsection:B.3}, we need to verify whether the output truly meets the six requirements. Therefore, we further perform a quality assessment of the Q\&A. The specific prompt is shown below(originally in Chinese, but shown here in English), and an example of the model's output is illustrated in Figure \ref{fig:example3}.
The \{Key points\} in the prompt refers to the output of section \ref{subsection:B.2} with examples shown in Figure \ref{fig:example1}. The {Questions and answers} in the prompt refer to the output of Section \ref{subsection:B.3}, with examples shown in Figure \ref{fig:example2}.

\begin{tcolorbox}[colback=gray!5,colframe=gray!80, breakable,title=Prompt]
\textbf{\# Task Description}
\newline You are an evaluation expert, skilled in assessing the rationality of a question and answer design based on certain criteria. A Q\&A consists of a question and four options.
\textbf{\newline\# Specific Steps}
\newline Analyze according to the following steps: Ensure that the question design effectively tests others' understanding of the content without being influenced by formal differences.
\newline[Consistency] All options should be approximately the same length to avoid obvious differences in length. Ensure all options maintain consistency in tone and professionalism, with similar wording styles to prevent identifying the correct answer through stylistic differences.
\newline[Distractibility] Incorrect options should be designed to be confusing and seemingly reasonable, making them not easy to eliminate through common sense. Ensure that incorrect options are somewhat persuasive, rather than just hypothetical or obviously wrong.
\newline[Avoid Image Element Misguiding] Ensure that all options mentioning image elements align with or are similar to the actual content, to avoid easily eliminating incorrect options due to errors in image details.
\newline[Prevent Keyword and Pattern Recognition] Avoid explicit keyword matches between the question and the options. Ensure that there is no direct verbal association between the way the question is asked and the correct answer, to prevent easy inference.
\newline[Unique Correct Answer] Among all the options, ensure that only one is the correct answer. Avoid ambiguity or vagueness that allows each option to have only one clear judgment.
\newline[Core Assessment] The design of questions and answers must revolve around the key information in the assessment points.
\newline[Comprehensive Judgment] Determine whether the Q\&A meets all the above requirements, and directly output yes/no without any additional characters.
\textbf{\newline\# Output Format }(Please strictly follow the format below)
\newline[Consistency] One sentence judging whether consistency is met and briefly explaining the reason.
\newline[Distractibility] One sentence judging whether confusion is met and briefly explaining the reason.
\newline[Avoid Image Element Misguiding] One sentence judging whether avoiding misleading image elements is met and briefly explaining the reason.
\newline[Prevent Keyword and Pattern Recognition] One sentence judging whether preventing keyword and pattern recognition is met and briefly explaining the reason.
\newline[Unique Correct Answer] One sentence judging whether only one correct answer is met and briefly explaining the reason.
\newline[Core Assessment] One sentence judging whether core assessment is met and briefly explaining the reason.
\newline[Comprehensive Judgment] Yes/No
\textbf{\newline\# Assessment Points}
\newline\{Key points\}
\textbf{\newline\# Q\&A}
\newline \{Questions and answers\}
\end{tcolorbox}
\subsection{Advanced Filtering}
After completing the data quality assessment in \ref{subsection:C.1}, to ensure visual dependency and prevent keyword and pattern recognition, we use the following prompt to filter the collected high-quality QA. If the model can correctly answer without accompanying images, the question is discarded and regenerated until true visual dependency is achieved.

\begin{tcolorbox}[colback=gray!5,colframe=gray!80, breakable,title=Prompt]
Question: \{Question\}
 \newline A. \{A\}
 \newline B. \{B\}
 \newline C. \{C\}
 \newline D. \{D\}
 \newline Directly output the number of the correct answer, do not output any other extra characters.
\end{tcolorbox}

\section{Large Vision Language Models}
% \begin{itemize}
%     \item \textbf{InternVL2}\cite{chen2024fargpt4vclosinggap}is an extension of InternLM2, using InternViT-6B as vision encoder, while also featuring MLP projector sandwiched between them.
%     \item \textbf{Qwen2-VL}\cite{wang2024qwen2vlenhancingvisionlanguagemodels}is an extension of Qwen2-7B, incorporating a vision encoder and a vision-language fusion module to enhance multi-modal capabilities.
%     \item \textbf{MiniCPM-V-2.6}\cite{yao2024minicpmvgpt4vlevelmllm} is also an extension of Qwen2-7B, using SigLip-400M as the vision encoder, and introducing a adapter between them.
%     \item \textbf{LLaVA-OneVision}\cite{li2024llavaonevisioneasyvisualtask} also employs SigLip as the vision encoder, selects Qwen-2 as the LLM, and uses a two-layer MLP to project image features into the word embedding space.
%     \item \textbf{DeepSeek-VL}\cite{lu2024deepseekvlrealworldvisionlanguageunderstanding} employs two different vision encoders and uses DeepSeek LLM as the language decoder, utilizing a two-layer MLP as adapter.
%     \item \textbf{GPT4o}\cite{achiam2023gpt} is an cutting-edge large multimodal model from OpenAI that builds on the success of previous versions to deliver even more accurate, coherent, and contextually aware text generation by leveraging a larger dataset and refined transformer architecture.
% \end{itemize}
\begin{itemize}
    \item \textbf{InternVL2} is an extension of InternLM2, using InternViT as vision encoder, while also featuring MLP projector sandwiched between them.
    \item \textbf{Qwen2-VL} is an extension of Qwen2-7B, incorporating a vision encoder and a vision-language fusion module to enhance multi-modal capabilities.
    \item \textbf{MiniCPM-V-2.6} is also an extension of Qwen2-7B, using SigLip-400M as the vision encoder, and introducing a adapter between them.
    \item \textbf{LLaVA-OneVision} also employs SigLip as the vision encoder, selects Qwen-2 as the LLM, and uses a two-layer MLP to project image features into the word embedding space.
    \item \textbf{DeepSeek-VL} employs two different vision encoders and uses DeepSeek LLM as the language decoder, utilizing a two-layer MLP as adapter.
    \item \textbf{GPT4o} is an cutting-edge large multimodal model from OpenAI that builds on the success of previous versions to deliver even more accurate, coherent, and contextually aware text generation by leveraging a larger dataset and refined transformer architecture.
\end{itemize}

\section{Model Hyper-parameter Details}
\begin{table}
\centering
\begin{tblr}{
  column{2} = {c},
  column{3} = {c},
  hline{1,16} = {-}{0.08em},
  hline{2} = {-}{0.05em},
}
 & Temperature & Top\_k\\
InternVL2-1B & 1.0 & 50\\
InternVL2-2B & 1.0 & 50\\
InternVL2-4B & 1.0 & 50\\
InternVL2-8B & 1.0 & 50\\
InternVL2-26B & 1.0 & 50\\
InternVL2-40B & 1.0 & 50\\
InternVL1.5-26B & 1.0 & 50\\
Qwen2-VL-2B-Instruct & 0.01 & 1\\
Qwen2-VL-7B-Instruct & 0.01 & 1\\
Qwen2-VL-72B-Instruct & 1.0 & 1\\
DeepSeek-VL-7B-chat & 1.0 & \\
llava-onevision-qwen2-7b & 0.7 & 20\\
llava-onevision-qwen2-0.5b & 0.7 & 20\\
MiniCPM-V 2.6 (8B) & 0.7 & 100
\end{tblr}
\caption{The hyper-parameter of all models evaluated in this work.}
\label{tab:param}
\end{table}
The specific parameters used by all models in this paper are shown in Table \ref{tab:param}.

\begin{figure}[t]
  \centering
  \includegraphics[width=1.0 \linewidth]{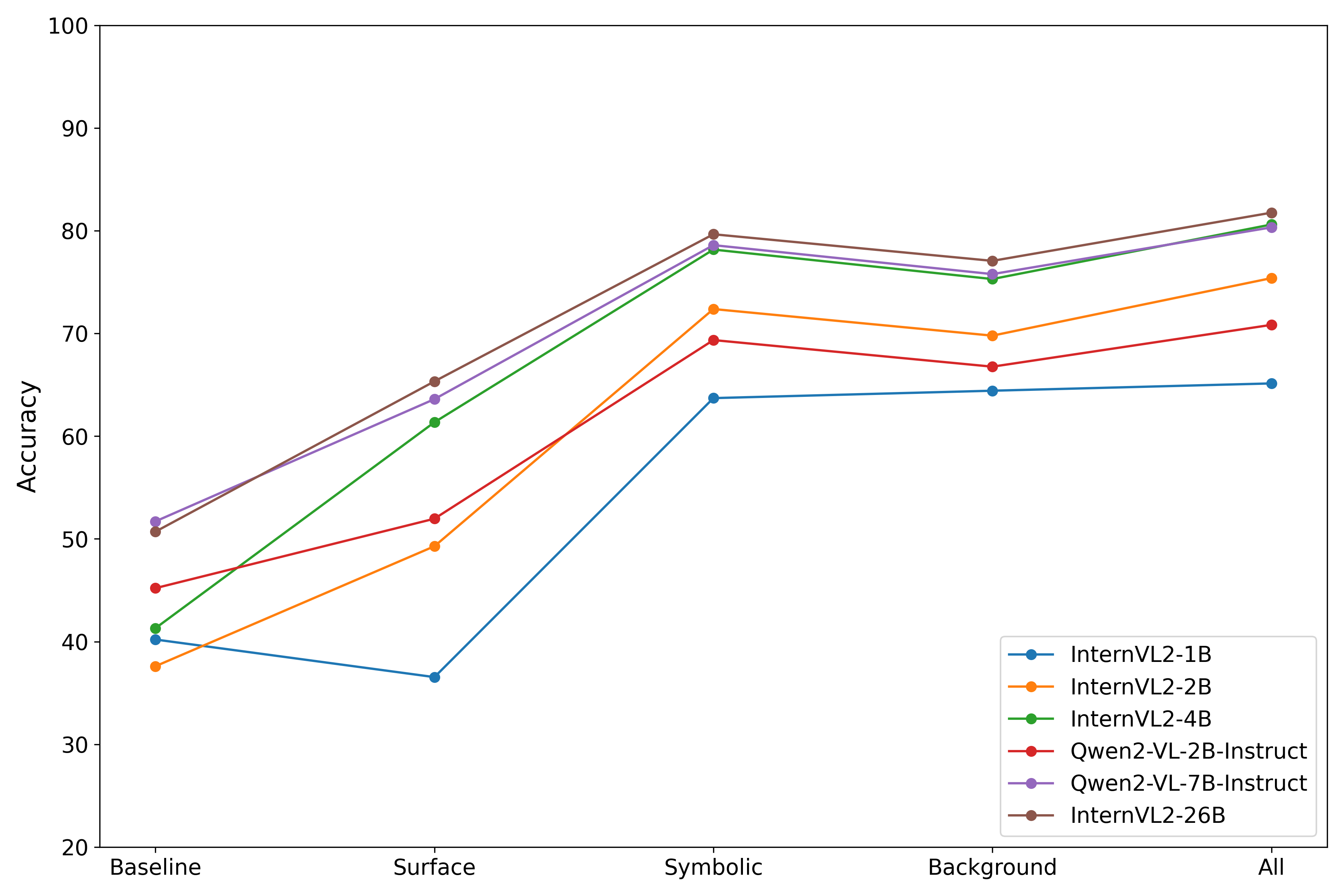}
   \caption{Impact of Key Points from Different Levels on Implicit Meaning Comprehension. Baseline indicates no additional information, Surface, Symbolic, and Background represent the injection of key points from Surface-level content, Symbolic meaning, and Background knowledge, respectively. All indicates the simultaneous injection of key points from all three levels.}
   \label{fig:infl}
\end{figure}

\section{Can Image Descriptions Help the Model Understand Implicit Meaning?}
To investigate the relationship between the key points of image descriptions at different levels (as shown in Figure \ref{fig:example1}) and the understanding of deep semantic meanings in images, we selected models with varying parameter scales from InternVL2 and Qwen2-VL for our study. This is because these two model frameworks are relatively classical and possess a broad range of parameter sizes, which makes them suitable for comprehensive analysis. We then evaluated the impact of providing supplementary information on the implicit meaning comprehension task by using key points extracted from Surface-level content, Symbolic meaning, and Background knowledge (as shown in the lower part of Figure \ref{fig:example1}) individually and in combination. The results are shown in Figure \ref{fig:infl}.

The results indicate that adding key points from the image descriptions of the three other levels significantly improves the model's accuracy in the implicit meaning comprehension task. Injecting Symbolic Meaning alone can enhance the performance of each model by approximately 20-30\%. When all three levels of key points are injected simultaneously, the accuracy of each model is further improved, with larger-scale models achieving an accuracy rate exceeding 80\%, which surpasses human performance. Therefore, we can hypothesize that when models receive information from various levels, they learn useful knowledge that aids in understanding implicit meaning. However, it is worth noting that due to the correlation between key points and the final QA, we cannot entirely rule out the possibility that some of the improvement is attributed to this correlation. Nonetheless, we can conclude that the injection of shallow-level information helps models understand implicit meaning.

\section{What Strategies Augment a Model's Comprehension of Implicit Meanings in Images?}
In order to enhance our model's ability to comprehend implicit meanings, we propose for the first time an innovative method involving the construction of multi-turn dialogues. This approach leverages both images with implicit meanings and artificially constructed virtual images (with virtual images being textual descriptions elaborating a specific image). Specifically, we employ Qwen2-VL-7B as the foundational model. For an image that conveys implicit meaning, our training data is structured into multi-turn dialogues composed of three modules.

The first module involves an input of the image paired with text, where the text comprises various questions, such as "How would you describe this image to a stranger?" This prompts the pre-trained Qwen2-VL-7B to generate a caption of the image. The text input and the caption output form the dialogue in this module, primarily serving the purpose of maintaining the original general capabilities of Qwen2-VL-7B.

In the second module, we utilize Qwen2-VL-72B and a large-scale model like GPT4o to construct multi-turn dialogues from three dimensions: Surface-level content, Symbolic meaning, and Background knowledge. This is designed to infuse the model with knowledge related to surface visual elements, symbolic meanings, and background information.

The third module focuses on extracting the reasoning processes associated with the Implicit Meaning from the images using Qwen2-VL-72B and GPT4o. It generates high-quality chain-of-thought (CoT) data. The CoT data and text inputs form the dialogue in this module, aimed at enhancing the model's inferential capabilities in understanding implicit meanings within a multimodal context.

Simultaneously, to address the insufficiency of real images with implicit meanings, we generate a substantial number of virtual images with implicit meanings via large language models (LLMs). These virtual images are used in place of real ones and are subjected to the same multi-turn dialogue construction for training the model. This approach effectively compensates for the lack of real images with implicit meanings. 

As evidenced in Table \ref{tab:new benchmark}, our proposed method enables the Qwen2-VL-7B model to achieve an accuracy of 62.5\% on the Implicit Meaning Comprehension Task, surpassing the 59.3\% accuracy of GPT4o and the 60.1\% accuracy of Qwen2-VL-72B-Instruct. This demonstrates the effectiveness and superiority of our approach.

\begin{table*}
\centering
\begin{tblr}{
  width = \linewidth,
  colspec = {Q[300]Q[94]Q[133]Q[92]Q[121]Q[63]Q[140]},
  column{even} = {c},
  column{3} = {c},
  column{5} = {c},
  column{7} = {c},
  hline{1,17} = {-}{0.08em},
  hline{2} = {-}{0.05em},
  hline{14} = {1}{l},
  hline{14} = {2-6}{},
  hline{14} = {7}{r},
  hline{15} = {1}{l},
  hline{15} = {2-6}{},
  hline{15} = {7}{r},
  hline{16} = {1}{l},
  hline{16} = {2-6}{},
  hline{16} = {7}{r},
  % hline{17} = {1}{l},
  % hline{17} = {2-6}{},
  % hline{17} = {7}{r},
  % hline{16-17} = {1}{l},
  % hline{16-17} = {2-6}{},
  % hline{16-17} = {7}{r},
  %hline{1,19} = {-}{0.08em},
  %hline{2} = {-}{0.05em},
  %hline{17-18} = {1}{l},
  %hline{17-18} = {2-6}{},
  %hline{17-18} = {7}{r},
}
\textbf{Model}                 & \textbf{\# Params} & \textbf{Surface} & \textbf{Symbolic} & \textbf{Background} & \textbf{Mean} & \textbf{Implicit} \\
InternVL2-Llama3-76B\cite{chen2024fargpt4vclosinggap}           & 76B                & 74.7                   & 71.1              & 75.4                 & 73.7          & 53.8                      \\
Qwen2-VL-72B-Instruct\cite{wang2024qwen2vlenhancingvisionlanguagemodels}            & 72B                & 79.3                   & 82.6     & 81.6        & 81.2 & 60.1             \\
InternVL2-40B\cite{chen2024fargpt4vclosinggap}                  & 40B                & 79.5                   & 79.8              & 80.7                 & 80.0          & 58.7                      \\
InternVL1.5-26B\cite{chen2024fargpt4vclosinggap}                & 26B                & 74.1                   & 70.5              & 74.4                 & 73.0          & 54.7                      \\
InternVL2-26B\cite{chen2024fargpt4vclosinggap}                  & 26B                & 75.2                   & 71.8              & 73.9                 & 73.6          & 50.7                      \\
InternVL2-8B\cite{chen2024fargpt4vclosinggap}                   & 8B                 & 70.7                   & 73.6              & 73.7                 & 72.7          & 46.5                      \\
MiniCPM-V-2\_6\cite{yao2024minicpmvgpt4vlevelmllm}                   & 8B                 & 74.0                   & 74.1              & 79.2                 & 75.8          & 50.0                      \\
Qwen2-VL-7B-Instruct\cite{wang2024qwen2vlenhancingvisionlanguagemodels}             & 7B                 & 75.1                   & 81.1              & 79.3                 & 78.5          & 51.7                      \\
llava-onevision-qwen2-7b\cite{li2024llavaonevisioneasyvisualtask}  & 7B                 & 74.2                   & 72.9              & 76.2                 & 74.4          & 50.0                      \\
v2\_deepseek-vl-7b-chat\cite{lu2024deepseekvlrealworldvisionlanguageunderstanding}        & 7B                 & 58.8                   & 57.3              & 65.6                 & 60.6          & 38.1                      \\
% InternVL2-4B\cite{chen2024fargpt4vclosinggap}                   & 4B                 & 63.3                   & 68.9              & 69.8                 & 67.3          & 41.3                      \\
% InternVL2-2B\cite{chen2024fargpt4vclosinggap}                   & 2B                 & 58.2                   & 56.0              & 64.4                 & 59.5          & 37.6                      \\
Qwen2-VL-2B-Instruct\cite{wang2024qwen2vlenhancingvisionlanguagemodels}           & 2B                 & 70.3                   & 73.8              & 74.5                 & 72.9          & 45.2                      \\
llava-onevision-qwen2-0.5b\cite{li2024llavaonevisioneasyvisualtask} & 0.5B                 & 44.4                   & 45.0              & 33.3                 & 40.9          & 23.2                      \\
% InternVL2-1B\cite{chen2024fargpt4vclosinggap}                   & 1B                 & 49.9                   & 56.5              & 57.9                 & 54.8          & 40.3                      \\
GPT4o                          & -                  & \textbf{82.0}          & 80.8              & 79.8                 & 80.9          & 59.3                      \\
Ours                          & 7B                  & 78.4          & \textbf{84.6}              & \textbf{83.9}                 & \textbf{82.3}          & \textbf{62.5}                      \\

Human                          & -                  & 98.0                   & 88.0              & 86.0                 & 90.7          & 74.0                      
\end{tblr}
\caption{The benchmark includes the average accuracy (in percentages (\%)) on four tasks. Surface, Symbolic, Background, and Implicit represent Surface-level Content Understanding Task, Symbolic Meaning Interpretation Task, Background Knowledge Comprehension Task, and Implicit Meaning Comprehension Task, respectively. The Mean represents the average accuracy of the first three tasks.}
\label{tab:new benchmark}
\end{table*}

%为了探究图片不同level的相关知识与理解图像深度语义之间的关系，我们选取InternVL, Qwen2-VL的不同参数规模的模型作为研究对象，因为InternVL2及Qwen2-VL-72B在implict meaning comprehension task上表现最佳，与GPT4o旗鼓相当。然后，我们将Surface-level content, symbolic meaning, Background Knowledge中提取出来的关键点（如图1下半部分所示）分别以及同时作为implict meaning comprehension task的补充信息来评估其影响。结果如图4所示。
%结果表明，在进行implict meaning comprehension task加入三个其他Level的Keypoint信息，可以有效提升模型的准确率，单独Symbolic Meaning的注入能给各模型带来约20%-30%的提升，而将三个Level的Keypoint信息同时注入时，各模型的准确率得到了进一步提升，对于参数规模较大的模型，其效果已经超过了Human74%的准确率，因此我们可以假定，模型获取到各Level的信息后，会学习到有用的知识，对理解Implicit meaning带来帮助，但值得一提的是，由于Keypoint和最后的QA中存在一定的相关性，我们无法排除一部分提高是由于这种相关性导致的。但从整体而言，我们还是可以得出浅层信息的注入可以帮助模型进行Implicit meaning的理解

% 为了探究图片的相关信息与理解图像深度语义之间的关系
% 其中baseline表示无任何额外信息，Surface，symbolic，background，分别表示单独注入Surface-level content, symbolic meaning, Background Knowledge的Key points，all表示将三个信息同时注入
%分类：GPT4o写了隐晦表达 可能的分类，爬的10w数据按这个分类去分，一部分无法分类，剩下的让gpt继续分类，以此循环

%% file: main.bbl
\begin{thebibliography}{43}
\providecommand{\natexlab}[1]{#1}
\providecommand{\url}[1]{\texttt{#1}}
\expandafter\ifx\csname urlstyle\endcsname\relax
  \providecommand{\doi}[1]{doi: #1}\else
  \providecommand{\doi}{doi: \begingroup \urlstyle{rm}\Url}\fi

\bibitem[Achiam et~al.(2023)Achiam, Adler, Agarwal, Ahmad, Akkaya, Aleman, Almeida, Altenschmidt, Altman, Anadkat, et~al.]{achiam2023gpt}
Josh Achiam, Steven Adler, Sandhini Agarwal, Lama Ahmad, Ilge Akkaya, Florencia~Leoni Aleman, Diogo Almeida, Janko Altenschmidt, Sam Altman, Shyamal Anadkat, et~al.
\newblock Gpt-4 technical report.
\newblock \emph{arXiv preprint arXiv:2303.08774}, 2023.

\bibitem[Alayrac et~al.(2022)Alayrac, Donahue, Luc, Miech, Barr, Hasson, Lenc, Mensch, Millican, Reynolds, et~al.]{alayrac2022flamingo}
Jean-Baptiste Alayrac, Jeff Donahue, Pauline Luc, Antoine Miech, Iain Barr, Yana Hasson, Karel Lenc, Arthur Mensch, Katherine Millican, Malcolm Reynolds, et~al.
\newblock Flamingo: a visual language model for few-shot learning.
\newblock \emph{Advances in neural information processing systems}, 35:\penalty0 23716--23736, 2022.

\bibitem[Anil et~al.(2023)Anil, Dai, Firat, Johnson, Lepikhin, Passos, Shakeri, Taropa, Bailey, Chen, et~al.]{anil2023palm}
Rohan Anil, Andrew~M Dai, Orhan Firat, Melvin Johnson, Dmitry Lepikhin, Alexandre Passos, Siamak Shakeri, Emanuel Taropa, Paige Bailey, Zhifeng Chen, et~al.
\newblock Palm 2 technical report.
\newblock \emph{arXiv preprint arXiv:2305.10403}, 2023.

\bibitem[Awadalla et~al.(2023)Awadalla, Gao, Gardner, Hessel, Hanafy, Zhu, Marathe, Bitton, Gadre, Sagawa, Jitsev, Kornblith, Koh, Ilharco, Wortsman, and Schmidt]{awadalla2023openflamingoopensourceframeworktraining}
Anas Awadalla, Irena Gao, Josh Gardner, Jack Hessel, Yusuf Hanafy, Wanrong Zhu, Kalyani Marathe, Yonatan Bitton, Samir Gadre, Shiori Sagawa, Jenia Jitsev, Simon Kornblith, Pang~Wei Koh, Gabriel Ilharco, Mitchell Wortsman, and Ludwig Schmidt.
\newblock Openflamingo: An open-source framework for training large autoregressive vision-language models, 2023.

\bibitem[Bai et~al.(2023)Bai, Bai, Yang, Wang, Tan, Wang, Lin, Zhou, and Zhou]{bai2023qwen}
Jinze Bai, Shuai Bai, Shusheng Yang, Shijie Wang, Sinan Tan, Peng Wang, Junyang Lin, Chang Zhou, and Jingren Zhou.
\newblock Qwen-vl: A versatile vision-language model for understanding, localization, text reading, and beyond.
\newblock \emph{arXiv preprint arXiv:2308.12966}, 1\penalty0 (2):\penalty0 3, 2023.

\bibitem[Cai et~al.(2019)Cai, Cai, and Wan]{cai2019multi}
Yitao Cai, Huiyu Cai, and Xiaojun Wan.
\newblock Multi-modal sarcasm detection in twitter with hierarchical fusion model.
\newblock In \emph{Proceedings of the 57th annual meeting of the association for computational linguistics}, pages 2506--2515, 2019.

\bibitem[Chen et~al.(2015)Chen, Fang, Lin, Vedantam, Gupta, Dollar, and Zitnick]{chen2015microsoftcococaptionsdata}
Xinlei Chen, Hao Fang, Tsung-Yi Lin, Ramakrishna Vedantam, Saurabh Gupta, Piotr Dollar, and C.~Lawrence Zitnick.
\newblock Microsoft coco captions: Data collection and evaluation server, 2015.

\bibitem[Chen et~al.(2024)Chen, Wang, Tian, Ye, Gao, Cui, Tong, Hu, Luo, Ma, Ma, Wang, Dong, Yan, Guo, He, Shi, Jin, Xu, Wang, Wei, Li, Zhang, Zhang, Cai, Wen, Yan, Dou, Lu, Zhu, Lu, Lin, Qiao, Dai, and Wang]{chen2024fargpt4vclosinggap}
Zhe Chen, Weiyun Wang, Hao Tian, Shenglong Ye, Zhangwei Gao, Erfei Cui, Wenwen Tong, Kongzhi Hu, Jiapeng Luo, Zheng Ma, Ji Ma, Jiaqi Wang, Xiaoyi Dong, Hang Yan, Hewei Guo, Conghui He, Botian Shi, Zhenjiang Jin, Chao Xu, Bin Wang, Xingjian Wei, Wei Li, Wenjian Zhang, Bo Zhang, Pinlong Cai, Licheng Wen, Xiangchao Yan, Min Dou, Lewei Lu, Xizhou Zhu, Tong Lu, Dahua Lin, Yu Qiao, Jifeng Dai, and Wenhai Wang.
\newblock How far are we to gpt-4v? closing the gap to commercial multimodal models with open-source suites, 2024.

\bibitem[Chow et~al.(2023)Chow, Tan, and Kan]{chow2023travlritdontbimodal}
Keng~Ji Chow, Samson Tan, and Min-Yen Kan.
\newblock Travlr: Now you see it, now you don't! a bimodal dataset for evaluating visio-linguistic reasoning, 2023.

\bibitem[Contributors(2023)]{contributorsopencompass}
Opencompass Contributors.
\newblock Opencompass: A universal evaluation platform for foundation models (2023).
\newblock \emph{URL https://github. com/open-compass/opencompass}, 2023.

\bibitem[Dai et~al.(2023)Dai, Li, Li, Tiong, Zhao, Wang, Li, Fung, and Hoi]{dai2023instructblipgeneralpurposevisionlanguagemodels}
Wenliang Dai, Junnan Li, Dongxu Li, Anthony Meng~Huat Tiong, Junqi Zhao, Weisheng Wang, Boyang Li, Pascale Fung, and Steven Hoi.
\newblock Instructblip: Towards general-purpose vision-language models with instruction tuning, 2023.

\bibitem[de~Wit and Wagemans(2012)]{DEWIT2012665}
L. de Wit and J. Wagemans.
\newblock Visual perception.
\newblock In \emph{Encyclopedia of Human Behavior (Second Edition)}, pages 665--671. Academic Press, San Diego, second edition edition, 2012.

\bibitem[Fu et~al.(2024)Fu, Chen, Shen, Qin, Zhang, Lin, Yang, Zheng, Li, Sun, Wu, and Ji]{fu2024mmecomprehensiveevaluationbenchmark}
Chaoyou Fu, Peixian Chen, Yunhang Shen, Yulei Qin, Mengdan Zhang, Xu Lin, Jinrui Yang, Xiawu Zheng, Ke Li, Xing Sun, Yunsheng Wu, and Rongrong Ji.
\newblock Mme: A comprehensive evaluation benchmark for multimodal large language models, 2024.

\bibitem[Garner(1987)]{garner1987metacognition}
Ruth Garner.
\newblock \emph{Metacognition and reading comprehension.}
\newblock Ablex Publishing, 1987.

\bibitem[Gordon et~al.(2019)Gordon, Hohwy, Davidson, van Boxtel, and Tsuchiya]{gordon2019intermodulation}
Noam Gordon, Jakob Hohwy, Matthew~James Davidson, Jeroen~JA van Boxtel, and Naotsugu Tsuchiya.
\newblock From intermodulation components to visual perception and cognition-a review.
\newblock \emph{NeuroImage}, 199:\penalty0 480--494, 2019.

\bibitem[Goyal et~al.(2017)Goyal, Khot, Summers-Stay, Batra, and Parikh]{Goyal_2017_CVPR}
Yash Goyal, Tejas Khot, Douglas Summers-Stay, Dhruv Batra, and Devi Parikh.
\newblock Making the v in vqa matter: Elevating the role of image understanding in visual question answering.
\newblock In \emph{Proceedings of the IEEE Conference on Computer Vision and Pattern Recognition (CVPR)}, 2017.

\bibitem[He et~al.(2024)He, Wu, Zhou, Xuan, Liu, Yang, Zhu, and Huang]{he2024cmmubenchmarkchinesemultimodal}
Zheqi He, Xinya Wu, Pengfei Zhou, Richeng Xuan, Guang Liu, Xi Yang, Qiannan Zhu, and Hua Huang.
\newblock Cmmu: A benchmark for chinese multi-modal multi-type question understanding and reasoning, 2024.

\bibitem[Hiippala et~al.(2020)Hiippala, Alikhani, Haverinen, Kalliokoski, Logacheva, Orekhova, Tuomainen, Stone, and Bateman]{Hiippala_2020}
Tuomo Hiippala, Malihe Alikhani, Jonas Haverinen, Timo Kalliokoski, Evanfiya Logacheva, Serafina Orekhova, Aino Tuomainen, Matthew Stone, and John~A. Bateman.
\newblock Ai2d-rst: a multimodal corpus of 1000 primary school science diagrams.
\newblock \emph{Language Resources and Evaluation}, 55\penalty0 (3):\penalty0 661–688, 2020.

\bibitem[Hudson and Manning(2019)]{Hudson_2019_CVPR}
Drew~A. Hudson and Christopher~D. Manning.
\newblock Gqa: A new dataset for real-world visual reasoning and compositional question answering.
\newblock In \emph{Proceedings of the IEEE/CVF Conference on Computer Vision and Pattern Recognition (CVPR)}, 2019.

\bibitem[Li et~al.(2023{\natexlab{a}})Li, Wang, Wang, Ge, Ge, and Shan]{li2023seedbenchbenchmarkingmultimodalllms}
Bohao Li, Rui Wang, Guangzhi Wang, Yuying Ge, Yixiao Ge, and Ying Shan.
\newblock Seed-bench: Benchmarking multimodal llms with generative comprehension, 2023{\natexlab{a}}.

\bibitem[Li et~al.(2024{\natexlab{a}})Li, Zhang, Zhang, Guo, Zhang, Li, Zhang, Liu, and Li]{li2024llava}
Bo Li, Kaichen Zhang, Hao Zhang, Dong Guo, Renrui Zhang, Feng Li, Yuanhan Zhang, Ziwei Liu, and Chunyuan Li.
\newblock Llava-next: Stronger llms supercharge multimodal capabilities in the wild, 2024{\natexlab{a}}.

\bibitem[Li et~al.(2024{\natexlab{b}})Li, Zhang, Guo, Zhang, Li, Zhang, Zhang, Zhang, Li, Liu, and Li]{li2024llavaonevisioneasyvisualtask}
Bo Li, Yuanhan Zhang, Dong Guo, Renrui Zhang, Feng Li, Hao Zhang, Kaichen Zhang, Peiyuan Zhang, Yanwei Li, Ziwei Liu, and Chunyuan Li.
\newblock Llava-onevision: Easy visual task transfer, 2024{\natexlab{b}}.

\bibitem[Li et~al.(2023{\natexlab{b}})Li, Li, Savarese, and Hoi]{li2023blip}
Junnan Li, Dongxu Li, Silvio Savarese, and Steven Hoi.
\newblock Blip-2: Bootstrapping language-image pre-training with frozen image encoders and large language models.
\newblock In \emph{International conference on machine learning}, pages 19730--19742. PMLR, 2023{\natexlab{b}}.

\bibitem[Liu et~al.(2024{\natexlab{a}})Liu, Li, Li, and Lee]{Liu_2024_CVPR}
Haotian Liu, Chunyuan Li, Yuheng Li, and Yong~Jae Lee.
\newblock Improved baselines with visual instruction tuning.
\newblock In \emph{Proceedings of the IEEE/CVF Conference on Computer Vision and Pattern Recognition (CVPR)}, pages 26296--26306, 2024{\natexlab{a}}.

\bibitem[Liu et~al.(2024{\natexlab{b}})Liu, Li, Wu, and Lee]{liu2024visual}
Haotian Liu, Chunyuan Li, Qingyang Wu, and Yong~Jae Lee.
\newblock Visual instruction tuning.
\newblock \emph{Advances in neural information processing systems}, 36, 2024{\natexlab{b}}.

\bibitem[Liu et~al.(2025)Liu, Duan, Zhang, Li, Zhang, Zhao, Yuan, Wang, He, Liu, et~al.]{liu2025mmbench}
Yuan Liu, Haodong Duan, Yuanhan Zhang, Bo Li, Songyang Zhang, Wangbo Zhao, Yike Yuan, Jiaqi Wang, Conghui He, Ziwei Liu, et~al.
\newblock Mmbench: Is your multi-modal model an all-around player?
\newblock In \emph{European Conference on Computer Vision}, pages 216--233. Springer, 2025.

\bibitem[Liu et~al.(2024{\natexlab{c}})Liu, Fang, Feng, Du, Zhang, Wang, Bai, Zhao, Fan, Gan, Lin, Li, Ni, Wu, Narsupalli, Zheng, Li, Hu, Xu, Chen, Yang, Liu, Liu, Huang, Zhang, and Ni]{liu2024iibenchimageimplicationunderstanding}
Ziqiang Liu, Feiteng Fang, Xi Feng, Xinrun Du, Chenhao Zhang, Zekun Wang, Yuelin Bai, Qixuan Zhao, Liyang Fan, Chengguang Gan, Hongquan Lin, Jiaming Li, Yuansheng Ni, Haihong Wu, Yaswanth Narsupalli, Zhigang Zheng, Chengming Li, Xiping Hu, Ruifeng Xu, Xiaojun Chen, Min Yang, Jiaheng Liu, Ruibo Liu, Wenhao Huang, Ge Zhang, and Shiwen Ni.
\newblock Ii-bench: An image implication understanding benchmark for multimodal large language models, 2024{\natexlab{c}}.

\bibitem[Lu et~al.(2024)Lu, Liu, Zhang, Wang, Dong, Liu, Sun, Ren, Li, Yang, Sun, Deng, Xu, Xie, and Ruan]{lu2024deepseekvlrealworldvisionlanguageunderstanding}
Haoyu Lu, Wen Liu, Bo Zhang, Bingxuan Wang, Kai Dong, Bo Liu, Jingxiang Sun, Tongzheng Ren, Zhuoshu Li, Hao Yang, Yaofeng Sun, Chengqi Deng, Hanwei Xu, Zhenda Xie, and Chong Ruan.
\newblock Deepseek-vl: Towards real-world vision-language understanding, 2024.

\bibitem[Machajdik and Hanbury(2010)]{machajdik2010affective}
Jana Machajdik and Allan Hanbury.
\newblock Affective image classification using features inspired by psychology and art theory.
\newblock In \emph{Proceedings of the 18th ACM international conference on Multimedia}, pages 83--92, 2010.

\bibitem[Movement(2010)]{cartoonmovement}
Cartoon Movement.
\newblock Cartoon movement website.
\newblock \url{https://www.cartoonmovement.com/search?query=&sort=created&order=desc}, 2010.
\newblock Accessed: 2023-10-05.

\bibitem[Radford et~al.(2021)Radford, Kim, Hallacy, Ramesh, Goh, Agarwal, Sastry, Askell, Mishkin, Clark, et~al.]{radford2021learning}
Alec Radford, Jong~Wook Kim, Chris Hallacy, Aditya Ramesh, Gabriel Goh, Sandhini Agarwal, Girish Sastry, Amanda Askell, Pamela Mishkin, Jack Clark, et~al.
\newblock Learning transferable visual models from natural language supervision.
\newblock In \emph{International conference on machine learning}, pages 8748--8763. PMLR, 2021.

\bibitem[Ray(2023)]{ray2023chatgpt}
Partha~Pratim Ray.
\newblock Chatgpt: A comprehensive review on background, applications, key challenges, bias, ethics, limitations and future scope.
\newblock \emph{Internet of Things and Cyber-Physical Systems}, 3:\penalty0 121--154, 2023.

\bibitem[Sun et~al.(2023)Sun, Fang, Wu, Wang, and Cao]{sun2023evaclipimprovedtrainingtechniques}
Quan Sun, Yuxin Fang, Ledell Wu, Xinlong Wang, and Yue Cao.
\newblock Eva-clip: Improved training techniques for clip at scale, 2023.

\bibitem[Sun et~al.(2024)Sun, Cui, Zhang, Zhang, Yu, Luo, Wang, Rao, Liu, Huang, and Wang]{sun2024generativemultimodalmodelsincontext}
Quan Sun, Yufeng Cui, Xiaosong Zhang, Fan Zhang, Qiying Yu, Zhengxiong Luo, Yueze Wang, Yongming Rao, Jingjing Liu, Tiejun Huang, and Xinlong Wang.
\newblock Generative multimodal models are in-context learners, 2024.

\bibitem[Tong et~al.(2024)Tong, Liu, Zhai, Ma, LeCun, and Xie]{tong2024eyes}
Shengbang Tong, Zhuang Liu, Yuexiang Zhai, Yi Ma, Yann LeCun, and Saining Xie.
\newblock Eyes wide shut? exploring the visual shortcomings of multimodal llms.
\newblock In \emph{Proceedings of the IEEE/CVF Conference on Computer Vision and Pattern Recognition}, pages 9568--9578, 2024.

\bibitem[Touvron et~al.(2023{\natexlab{a}})Touvron, Lavril, Izacard, Martinet, Lachaux, Lacroix, Rozière, Goyal, Hambro, Azhar, Rodriguez, Joulin, Grave, and Lample]{touvron2023llamaopenefficientfoundation}
Hugo Touvron, Thibaut Lavril, Gautier Izacard, Xavier Martinet, Marie-Anne Lachaux, Timothée Lacroix, Baptiste Rozière, Naman Goyal, Eric Hambro, Faisal Azhar, Aurelien Rodriguez, Armand Joulin, Edouard Grave, and Guillaume Lample.
\newblock Llama: Open and efficient foundation language models, 2023{\natexlab{a}}.

\bibitem[Touvron et~al.(2023{\natexlab{b}})Touvron, Martin, Stone, Albert, Almahairi, Babaei, Bashlykov, Batra, Bhargava, Bhosale, Bikel, Blecher, Ferrer, Chen, Cucurull, Esiobu, Fernandes, Fu, Fu, Fuller, Gao, Goswami, Goyal, Hartshorn, Hosseini, Hou, Inan, Kardas, Kerkez, Khabsa, Kloumann, Korenev, Koura, Lachaux, Lavril, Lee, Liskovich, Lu, Mao, Martinet, Mihaylov, Mishra, Molybog, Nie, Poulton, Reizenstein, Rungta, Saladi, Schelten, Silva, Smith, Subramanian, Tan, Tang, Taylor, Williams, Kuan, Xu, Yan, Zarov, Zhang, Fan, Kambadur, Narang, Rodriguez, Stojnic, Edunov, and Scialom]{touvron2023llama2openfoundation}
Hugo Touvron, Louis Martin, Kevin Stone, Peter Albert, Amjad Almahairi, Yasmine Babaei, Nikolay Bashlykov, Soumya Batra, Prajjwal Bhargava, Shruti Bhosale, Dan Bikel, Lukas Blecher, Cristian~Canton Ferrer, Moya Chen, Guillem Cucurull, David Esiobu, Jude Fernandes, Jeremy Fu, Wenyin Fu, Brian Fuller, Cynthia Gao, Vedanuj Goswami, Naman Goyal, Anthony Hartshorn, Saghar Hosseini, Rui Hou, Hakan Inan, Marcin Kardas, Viktor Kerkez, Madian Khabsa, Isabel Kloumann, Artem Korenev, Punit~Singh Koura, Marie-Anne Lachaux, Thibaut Lavril, Jenya Lee, Diana Liskovich, Yinghai Lu, Yuning Mao, Xavier Martinet, Todor Mihaylov, Pushkar Mishra, Igor Molybog, Yixin Nie, Andrew Poulton, Jeremy Reizenstein, Rashi Rungta, Kalyan Saladi, Alan Schelten, Ruan Silva, Eric~Michael Smith, Ranjan Subramanian, Xiaoqing~Ellen Tan, Binh Tang, Ross Taylor, Adina Williams, Jian~Xiang Kuan, Puxin Xu, Zheng Yan, Iliyan Zarov, Yuchen Zhang, Angela Fan, Melanie Kambadur, Sharan Narang, Aurelien Rodriguez, Robert Stojnic, Sergey Edunov, and Thomas
  Scialom.
\newblock Llama 2: Open foundation and fine-tuned chat models, 2023{\natexlab{b}}.

\bibitem[Wang et~al.(2024{\natexlab{a}})Wang, Bai, Tan, Wang, Fan, Bai, Chen, Liu, Wang, Ge, Fan, Dang, Du, Ren, Men, Liu, Zhou, Zhou, and Lin]{wang2024qwen2vlenhancingvisionlanguagemodels}
Peng Wang, Shuai Bai, Sinan Tan, Shijie Wang, Zhihao Fan, Jinze Bai, Keqin Chen, Xuejing Liu, Jialin Wang, Wenbin Ge, Yang Fan, Kai Dang, Mengfei Du, Xuancheng Ren, Rui Men, Dayiheng Liu, Chang Zhou, Jingren Zhou, and Junyang Lin.
\newblock Qwen2-vl: Enhancing vision-language model's perception of the world at any resolution, 2024{\natexlab{a}}.

\bibitem[Wang et~al.(2024{\natexlab{b}})Wang, Chen, Zhu, Luo, Li, Yan, Zhang, Huang, Sun, and Liu]{wang2024browseconcentratecomprehendingmultimodal}
Ziyue Wang, Chi Chen, Yiqi Zhu, Fuwen Luo, Peng Li, Ming Yan, Ji Zhang, Fei Huang, Maosong Sun, and Yang Liu.
\newblock Browse and concentrate: Comprehending multimodal content via prior-llm context fusion, 2024{\natexlab{b}}.

\bibitem[Yang et~al.(2024)Yang, Li, Dong, Xia, and Sui]{yang2024largemultimodalmodelsuncover}
Yixin Yang, Zheng Li, Qingxiu Dong, Heming Xia, and Zhifang Sui.
\newblock Can large multimodal models uncover deep semantics behind images?, 2024.

\bibitem[Yao et~al.(2024)Yao, Yu, Zhang, Wang, Cui, Zhu, Cai, Li, Zhao, He, Chen, Zhou, Zou, Zhang, Hu, Zheng, Zhou, Cai, Han, Zeng, Li, Liu, and Sun]{yao2024minicpmvgpt4vlevelmllm}
Yuan Yao, Tianyu Yu, Ao Zhang, Chongyi Wang, Junbo Cui, Hongji Zhu, Tianchi Cai, Haoyu Li, Weilin Zhao, Zhihui He, Qianyu Chen, Huarong Zhou, Zhensheng Zou, Haoye Zhang, Shengding Hu, Zhi Zheng, Jie Zhou, Jie Cai, Xu Han, Guoyang Zeng, Dahai Li, Zhiyuan Liu, and Maosong Sun.
\newblock Minicpm-v: A gpt-4v level mllm on your phone, 2024.

\bibitem[You et~al.(2023)You, Zhang, Gan, Du, Zhang, Wang, Cao, Chang, and Yang]{you2023ferretrefergroundgranularity}
Haoxuan You, Haotian Zhang, Zhe Gan, Xianzhi Du, Bowen Zhang, Zirui Wang, Liangliang Cao, Shih-Fu Chang, and Yinfei Yang.
\newblock Ferret: Refer and ground anything anywhere at any granularity, 2023.

\bibitem[Yue et~al.(2024)Yue, Ni, Zhang, Zheng, Liu, Zhang, Stevens, Jiang, Ren, Sun, Wei, Yu, Yuan, Sun, Yin, Zheng, Yang, Liu, Huang, Sun, Su, and Chen]{Yue_2024_CVPR}
Xiang Yue, Yuansheng Ni, Kai Zhang, Tianyu Zheng, Ruoqi Liu, Ge Zhang, Samuel Stevens, Dongfu Jiang, Weiming Ren, Yuxuan Sun, Cong Wei, Botao Yu, Ruibin Yuan, Renliang Sun, Ming Yin, Boyuan Zheng, Zhenzhu Yang, Yibo Liu, Wenhao Huang, Huan Sun, Yu Su, and Wenhu Chen.
\newblock Mmmu: A massive multi-discipline multimodal understanding and reasoning benchmark for expert agi.
\newblock In \emph{Proceedings of the IEEE/CVF Conference on Computer Vision and Pattern Recognition (CVPR)}, pages 9556--9567, 2024.

\end{thebibliography}
